\documentclass[runningheads]{llncs}

\PassOptionsToPackage{table}{xcolor}

\usepackage{eccv}

\usepackage{eccvabbrv}

\usepackage{graphicx}
\usepackage{booktabs}
\usepackage{makecell}
\usepackage{enumitem}

\usepackage{changepage}

\usepackage{colortbl}
\usepackage{xcolor}

\usepackage{adjustbox}

\usepackage{multirow}
\usepackage{rotating}
\usepackage{siunitx}
\usepackage{tabularray}

\usepackage{listings}
\usepackage{microtype}
\microtypesetup{expansion=false}

\newsavebox\CBox

\definecolor{gold}{RGB}{255, 215, 0}
\definecolor{silver}{RGB}{192, 192, 192}
\definecolor{bronze}{RGB}{205, 127, 50}

\definecolor{lightred}{HTML}{FF9999}
\definecolor{lightyellow}{HTML}{FFFF99}
\definecolor{lightorange}{HTML}{FFCC99}

\definecolor{DiffGreen}{HTML}{16A34A}
\definecolor{DiffRed}{HTML}{DC2626}
\definecolor{DiffGray}{HTML}{6B7280}

\usepackage{xspace}

\newcommand{\covopt}{PRAS\xspace}
\newcommand{\covoptLong}{Parallel Representative Assignment Smoothing\xspace}
\newcommand{\ourCompaction}{POPSpa\xspace}
\newcommand{\ourCompression}{SOG-XT\xspace}

\newcommand{\MipNeRF}{\mbox{\textit{Mip-NeRF~360}}\xspace}
\newcommand{\TandT}{\textit{Tanks and Temples}\xspace}
\newcommand{\DeepBlending}{\textit{Deep~Blending}\xspace}
\newcommand{\SyntheticNeRF}{\textit{Synthetic~NeRF}\xspace}

\newcommand{\venuetag}[1]{%
  \begingroup
  \setlength{\fboxsep}{0.6pt}%
  \textsf{\scriptsize\strut\,#1\,}%
  \endgroup
}

\definecolor{codebg}{RGB}{248,248,248}
\definecolor{codekw}{RGB}{0,64,128}
\definecolor{codecm}{RGB}{96,96,96}
\definecolor{codestr}{RGB}{128,0,0}
\definecolor{codeln}{RGB}{120,120,120}

\lstdefinestyle{pyreadable}{
  language=Python,
  backgroundcolor=\color{codebg},
  basicstyle=\ttfamily\scriptsize,
  keywordstyle=\color{codekw}\bfseries,
  commentstyle=\color{codecm},
  stringstyle=\color{codestr},
  showstringspaces=false,
  columns=fullflexible,
  keepspaces=true,
  tabsize=4,
  breaklines=true,
  breakatwhitespace=false,
  numbers=none,
  frame=single,
  framerule=0.2pt,
  rulecolor=\color{codeln},
  aboveskip=1.5mm,
  belowskip=1.5mm,
}

\usepackage[accsupp]{axessibility}  %

\usepackage{hyperref}

\usepackage{orcidlink}

\begin{document}

\title{KISS-GS: 3D Gaussian Splatting Compression Kept Simple}

\titlerunning{KISS-GS}

\author{
Wieland Morgenstern\inst{1,2}\orcidlink{0000-0001-5817-7464} \and
Friedrich Elias Branschke\inst{1,3}\orcidlink{0009-0002-5599-7445} \and
\\
Florian Fleischmann\inst{1,3}\orcidlink{0009-0004-5039-3305} \and
Adrian Szatmari\inst{1,3}\orcidlink{0009-0000-8814-0566}\and
Paul Schlack\inst{1}\orcidlink{0009-0006-9471-2878} \and
\\
Florian Barthel\inst{1,2}\orcidlink{0009-0004-7264-1672} \and
Peter Eisert\inst{1,2}\orcidlink{0000-0001-8378-4805} \and
Anna Hilsmann\inst{1,2}\orcidlink{0000-0002-2086-0951}
}

\authorrunning{W.~Morgenstern et al.}

\institute{Fraunhofer HHI, Germany\\
\email{\{first\}.\{last\}@hhi.fraunhofer.de}\and
Humboldt-Universität zu Berlin, Germany \and
Technische Universität Berlin, Germany
}

\maketitle

\begin{abstract}
Scene reconstruction with 3D Gaussian Splatting (3DGS) has become common, however deployment remains painful as the uncompressed file sizes can be massive. Current 3DGS compression systems combine multiple strategies for file size reduction, which can obscure where gains come from and limit component reuse across training pipelines. To make the gains more transparent, we propose KISS-GS, a modular compression pipeline named after the principle of keeping things simple, designed to decouple compression entirely from training.
Given a 3DGS scene reconstructed with vanilla 3DGS, we are able to reduce it through compaction by 15.7x using a combination of state-of-the-art pruning schemes. Then we encode it into an image-based format designed for simple, ubiquitous decoding.
With the SOG-XT format, we propose a novel extension to Self-Organizing Gaussians with two main contributions: (i) Self-organizing 2D Codebooks and (ii) Parallel Representative Assignment Smoothing (PRAS), which leverages the symmetry of quaternion and scale parameterizations to produce 2D attribute grids more amenable to encoding.
This encoding reduces scene size by 6.6x. We show that optional encoding-aware fine-tuning yields a further 2.2x. Across standard 3DGS benchmarks, our simple and modular approach thus achieves a total of 85× to 319× reductions in the size of the scene over uncompressed vanilla 3DGS, setting new benchmarks for real-world scenes and surpassing tightly integrated methods in rate-distortion. Decoding relies solely on web-native image formats, and the modular design makes each stage easy to combine with future advances in reconstruction and compaction.

Code and project page: \url{https://fraunhoferhhi.github.io/KISS-GS/}

\end{abstract}
    
\section{Introduction}
\label{sec:intro}

\begin{quote}\small\itshape
“Simplicity is a great virtue but it requires hard work to achieve it and education to appreciate it. And to make matters worse: complexity sells better.”

\hfill --- Edsger~W.~Dijkstra (1984), \emph{On the nature of Computing Science}~\cite{dijkstra1984ewd896}
\end{quote}

\begin{figure}[tb]
  \centering
  \includegraphics[width=\linewidth]{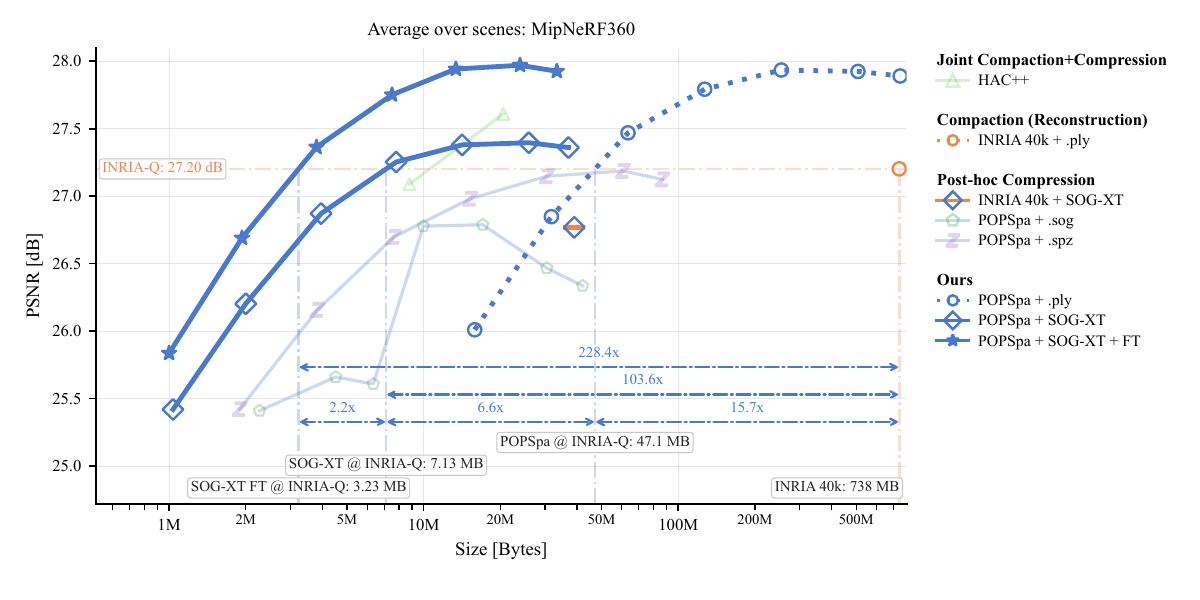}
  \caption{
  Rate distortion on \MipNeRF for KISS-GS, with quality measured by peak signal-to-noise ratio (PSNR).
At the INRIA 3DGS quality reference (INRIA-Q), KISS-GS reduces file size by $228\times$ from $738$ MB to $3.2$ MB.
  From the uncompressed INRIA \texttt{.ply} baseline, \ourCompaction matches INRIA-Q with a $15.7\times$ reduction, our \ourCompression encoding adds $6.6\times$, and encoding-
  aware fine-tuning adds a further $2.2\times$, outperforming HAC++\cite{chen2025hac-plus}.
}
  \label{fig:main_rd}
\end{figure}

\noindent
In practice, 3D Gaussian Splatting (3DGS) scenes are still commonly exchanged as raw \texttt{.ply} exports. 
These files are easy to produce, but large and inefficient for storage, streaming, and deployment. As 3DGS moves beyond research prototypes into hardware-constrained web viewers, mobile renderers, and AR/VR applications, file size, decoding cost, and memory footprint become obstacles.

Storage efficiency was not a focus of the original 3DGS formulation \cite{kerbl3Dgaussians}, leading to a rapidly growing body of work on compression.
The recent 3DGS.zip survey~\cite{3DGSzip2025} distinguishes two storage-reduction strategies, \emph{compaction}, which reduces the number of primitives, and \emph{compression}, which reduces the number of bits per primitive.
To avoid overloading the term \emph{compression}, we refer to this bits-per-primitive component as \emph{encoding}.
We add \emph{adaptation} as a third strategy, which uses encoding-aware training or fine-tuning to steer parameters toward configurations that are friendly to quantization and coding.
KISS-GS turns these three strategies into explicit pipeline stages, whose separate rate-distortion gains are previewed in \cref{fig:main_rd}.
Many methods combine multiple strategies to achieve their final rate-distortion scores.
While effective, this obscures where gains actually originate, creating an \emph{attribution gap}.
Without clear attribution, it is difficult to compare methods fairly and reuse components across pipelines.
Compaction alone can already yield large gains. Densification and pruning improvements can reduce primitive counts by an order of magnitude in some settings without large drops in reconstruction quality, as demonstrated by Mini-Splatting \cite{fang2024minisplatting, fang2024minisplattingv2}.
Some approaches, such as anchor-based families of methods, restructure the representation so fundamentally that compression becomes inseparable from the reconstruction regime and data format.
This \emph{training-format coupling} effectively locks the codec to a particular 3DGS training pipeline: when new reconstruction methods or parameterizations emerge, their improvements cannot be adopted without reworking the compression scheme itself.

We call our approach \emph{KISS-GS}, inspired by the KISS principle (``keep it simple, stupid'' \cite{rich1995clarence}) and restrict ourselves to a deliberately simple, browser-compatible decoding pipeline backed by widely supported image codecs.

Guided by this decoder-first stance, we ask three research questions (RQs):
\begin{enumerate}
    \item (\textbf{RQ1}) Can we close the \textbf{attribution gap} by modularizing the compression pipeline and measuring the contribution of each stage in turn?
    \item (\textbf{RQ2}) Can a \textbf{simple image-based encoding format} match more complex, integrated methods like HAC++\cite{chen2025hac-plus} across real-world 3DGS compression benchmarks?
    \item (\textbf{RQ3}) Does \textbf{encoding-aware fine-tuning} still deliver gains when constrained to a simple, web-friendly codec?
\end{enumerate}

To close the \textbf{attribution gap} (RQ1), we propose KISS-GS, a modular compression framework that separates compaction, encoding, and adaptation into distinct stages, enabling each stage to be replaced, ablated, or analyzed independently.
We instantiate the compaction stage with \ourCompaction, a reconstruction-independent module that combines state-of-the-art pruning schemes after the initial reconstruction, ensuring that gains from fewer primitives are never conflated with gains from better encoding.
To test whether a \textbf{simple image-based encoding format} can match more integrated methods (RQ2), we introduce \ourCompression, an image-based codec and storage format designed for web-friendly decoding.
We extend the Self-Organizing Gaussians (SOG)\cite{morgenstern2024compact} approach with (i) Self-organizing 2D codebooks for spherical harmonics and (ii) \covoptLong (\covopt{}) to improve compression rates while keeping the decoder simple.
To isolate \textbf{encoding-aware fine-tuning} under the constrained decoder model (RQ3), we systematically evaluate it with encoding-in-the-loop as a separate stage.

\section{Related work}
\label{sec:sota}
The original 3DGS paper \cite{kerbl3Dgaussians} focused on reconstruction quality and real-time rendering performance, without explicitly addressing storage or memory constraints. Due to its strong reconstruction quality with explicit primitives (in contrast to the implicit nature of NeRF-based approaches \cite{mildenhall2020nerf, mueller2022instant}), the community quickly adopted 3DGS. Many follow-up works have since investigated storage reduction, memory efficiency, and rendering performance, introducing modifications at different pipeline stages which can broadly be categorized into \textit{compaction}, \textit{encoding}, and \textit{adaptation}\cite{3DGSzip2025}. 

\textbf{Compaction} reduces the number of active Gaussians while preserving or improving reconstruction quality, directly impacting storage footprint and rendering cost. Many methods perform sensitivity-based pruning after or during training by estimating the importance of individual Gaussians. Various criteria have been proposed, including strict primitive budgets via score-based sampling \cite{taming20243dgs}, reconstruction error \cite{PapantonakisReduced3dgs, HansonSpeedy, HansonTuPUP3DGS, lee2026gaussianpop}, as well as primitive transmittance, opacity, or rendering contribution \cite{girish2024eagles, liu2024hemgs, lee2024compact}. Another class of methods controls the Gaussian population through sampling-based strategies. For example, MCMC-GS \cite{kheradmand2024mcmc} removes low-opacity primitives and relocates them to high-opacity regions to improve reconstruction under a fixed primitive budget.
Finally, some approaches formulate compaction as an explicit sparsity-constrained optimization problem: RDO-GS \cite{wang2024end} learns binary pruning masks jointly with the reconstruction objective using sparsity-inducing regularization, whereas GaussianSpa \cite{zhang2025gaussianspa} introduces an auxiliary sparsity variable and uses an augmented-Lagrangian formulation that alternates between standard reconstruction optimization and a projection step that enforces a target sparsity level.

\textbf{Encoding} reduces the number of bits required to represent each primitive without necessarily changing the primitive count. A common strategy is numerical compression of Gaussian attributes, often using vector quantization (VQ), particularly for view-dependent spherical harmonics \cite{navaneet2024compgs, navaneet2023compact3d, niedermayr2024compressed}. These approaches reduce bitwidth without altering primitive layout through quantization and entropy coding. 

Another class of methods replaces explicit attribute storage with structured representations that exploit spatial correlations between primitives. These include anchors, codebooks, multi-layer perceptrons (MLPs), octrees, triplanes, or hash grids to encode Gaussian attributes \cite{lu2024scaffold, chen2024hac, wu2024implicit, ren2024octreegs, lee2025compression3dgaussiansplatting}. \mbox{Scaffold-GS}~\cite{lu2024scaffold}, for example, predicts attributes from neighboring anchors via a lightweight MLP. Extensions include learning sparse primary and dense secondary anchors \cite{liu2024compgs}, hierarchical pyramids where coarser anchors predict finer ones~\cite{wang2024contextgs}, and querying hash-grid features at anchor locations \cite{chen2024hac, chen2025hac-plus}.

Another direction organizes Gaussian attributes into \textbf{structured 2D layouts} to leverage standard image codecs. Mapping 3D primitives to 2D grids via Morton curves or Self-Organizing Gaussians (SOG) \cite{morgenstern2024compact} produces locally smooth attribute maps optimized for conventional encoders. Although encoding requires an initial sorting step, decoding reduces to standard image decompression, which is widely hardware-accelerated and readily available in browser and mobile environments. This decoding simplicity has enabled \textbf{practical deployment} on mobile devices, AR/VR platforms, and browser-based viewers, leading to formalization of the \texttt{.sog} format \cite{playcanvas_sog}, which combines Self-Organizing Gaussians with VQ strategies for spherical harmonics. The format has been adopted in research and practical web pipelines \cite{playcanvas_engine, sparkjs, kellogg2023threejsgs}.
Alternative formats like \texttt{.spz} \cite{niantic2024spz} use structured quantization with a custom binary format.

\textbf{Adaptation} modifies scene parameters to make them more compatible with the target encoding. This can be achieved either during training or through post-training fine-tuning.
RDO-GS \cite{wang2024end} adds VQ-loss and rate-loss to the objective function, FCGS \cite{chen2025fcgs} incorporates entropy loss accounting for bit count, while HAC \cite{chen2024hac} and HAC++ \cite{chen2025hac-plus} combine rendering fidelity, entropy, masking, and hash-grid size losses.

\section{Modules of the KISS-GS pipeline}
\begin{figure}[t]
    \centering
    \includegraphics[width=\linewidth]{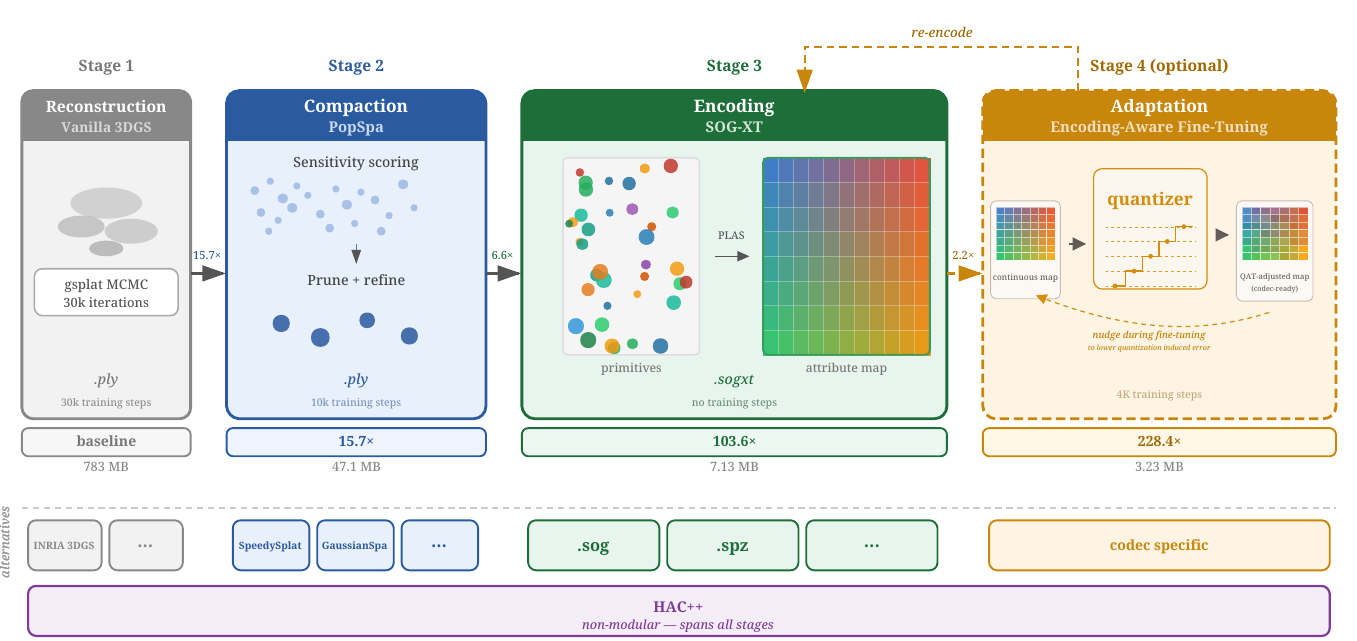}
\caption{The KISS-GS pipeline with four distinct stages. (1)~A standard 3DGS reconstruction yields a \texttt{.ply}, with primitive count setting the rate-distortion point. (2)~POPSpa compacts the scene. (3)~SOG-XT encodes the compacted primitives into attribute images. (4)~Optional codec-aware fine-tuning improves rate-distortion without changing the output format. Below the dashed separator, we show per-stage alternatives to the KISS-GS stages and contrast them with integrated methods like HAC++ spanning all stages as a single non-modular unit.}
    \label{fig:kiss_pipeline}
\end{figure}

\Cref{fig:kiss_pipeline} shows how the modular KISS-GS pipeline takes any off-the-shelf 3DGS scene as input, applies compaction, encodes the result with \ourCompression, and optionally fine-tunes it with the codec in the loop.
When competing methods differ in both file size and reconstruction quality, individual operating points listed in tables are difficult to compare.
We therefore use rate-distortion curves to compare this tradeoff across operating points.
When one curve dominates another in the Pareto sense (higher quality at the same file size, or smaller file size at the same quality) the dominant method is strictly better at every operating point.
We control the rate-distortion tradeoff primarily through primitive count, which has the additional benefit of reducing rendering cost and the work required for encoding and decoding. Using more primitives with heavier compression to reach the same operating point would instead increase all three, with no gain in rate or quality.
We describe each stage in turn.

\subsection{3D Reconstruction}
KISS-GS accepts any standard 3DGS \texttt{.ply} file as input, regardless of which implementation produced it.
This is a deliberate design choice: by decoupling compression from reconstruction, we avoid training-format coupling and ensure that improvements in reconstruction quality can be adopted without changes to the compression pipeline.
The standard 3DGS parameterization (means, covariances represented as quaternions and scales, spherical harmonic colors up to degree 3, and scalar opacity) is shared across all major implementations, and KISS-GS preserves this representation throughout.
Importantly, this parameterization relies on no MLPs, hash grids, or anchor structures, keeping the decoder simple.
The original training pipelines apply non-linear mappings to scales and opacities to constrain parameter ranges.
We use the same mappings because the resulting values are well suited to uniform quantization.

The 3DGS reconstruction ecosystem is actively evolving, with implementations varying in speed, quality, and license, including INRIA~\cite{kerbl3Dgaussians}, gsplat~\cite{ye2025gsplat}, Lichtfeld Studio~\cite{lichtfeld2025}, and Hahlbohm et al.\cite{hahlbohm2026fastergsanalyzingimprovinggaussian}.
Being decoupled from any specific implementation means KISS-GS can benefit from future reconstruction advances without modification.
Our main experiments use the gsplat MCMC reconstruction~\cite{kheradmand2024mcmc} with default parameters for 30k iterations; compatibility with alternative implementations is demonstrated in \cref{sec:eval}.

\subsection{Compaction with \ourCompaction}
\label{sec:compaction}
Densely trained 3DGS scenes typically contain substantial redundancy.
Since storage scales approximately linearly with the primitive count, compaction is essential to achieve high compression ratios.
We instantiate this stage with \emph{\ourCompaction}, a reconstruction-independent compaction recipe that starts from a trained 3DGS scene and combines score-based pruning from GaussianPOP~\cite{lee2026gaussianpop} with the alternating optimize-sparsify scheme of GaussianSpa \cite{zhang2025gaussianspa}, augmented by effective-rank regularization.
Its role is not to introduce a new pruning objective, but to provide a strong, replaceable module whose primitive-count gains can be measured separately from encoding gains.

\paragraph{Effective Rank Regularization.}
\label{sec:erank}
Although all covariance matrices maintain a full rank of 3, they exhibit widely different eigenvalue spectra, dictating the shape of the Gaussians. 
To quantify these shapes during training, we utilize the effective rank (erank) as a differentiable relaxation that measures the dimensions effectively occupied by the distribution.
Formally, for a 3D Gaussian with ordered scale attributes $s_1 \geq s_2 \geq s_3$, the normalized eigenvalues of its covariance matrix $\Sigma$ are defined as $p_i = \frac{s_i^2}{\sum_{j=1}^3 s_j^2}$.
The effective rank is formulated as the exponential of the Shannon entropy of the probability distribution induced by the eigenvalues of the covariance matrix: $\text{erank}(\Sigma)=\exp\left(- \sum_{i=1}^3  p_i \cdot \log p_i \right)$ \cite{roy_effective_2007}.
We employ the effective rank regularization (\cref{eq:erank_reg}) proposed by Hyung \etal \cite{hyung2024effective} to discourage the formation of extremely needle-shaped Gaussians, characterized by $\text{erank}(\Sigma_k)\approx 1$, which can overfit training views and produce artifacts from novel viewpoints. At the same time, this regularization favors disk-shaped Gaussians exhibiting an $\text{erank}(\Sigma_k)\approx 2$, which are better suited for surface reconstruction.

\begin{equation}
    \mathcal L_{\text{erank}} = \sum\limits_{k} \lambda_{\text{erank}}\max(-\log(\text{erank}(\Sigma_k)-1+\epsilon),0) + s_{3}
    \label{eq:erank_reg}
\end{equation}

\paragraph{First Pruning Stage.}
For each Gaussian primitive $k$, we use the error criterion of GaussianPOP~\cite{lee2026gaussianpop} to estimate how much the rendered image changes when $k$ is removed from the scene representation.
Let $C_\text{render}$ be the original rendered color and $C^\prime_\text{render}$ the color obtained after omitting Gaussian $k$; the per-primitive error $\Delta SE_k = \|C_\text{render}-C^\prime_\text{render}\|^2$ then quantifies the change in pixel colors, accumulated over all training views.
We compute $\Delta SE_k$ for all primitives using the efficient single-pass algorithm of~\cite{lee2026gaussianpop} and prune a fixed fraction with the lowest scores. 

\paragraph{Optimize-Sparsify Refinement.}
To counteract potential degradation caused by large pruning ratios, we refine the remaining primitives using the alternating optimize-sparsify scheme of GaussianSpa~\cite{zhang2025gaussianspa}.
GaussianSpa formulates primitive selection as a constrained optimization problem with an $\ell_0$ sparsity objective on primitive opacities:
\begin{equation}
  \min_{\mathbf{a},\mathbf{\Theta}}\ \mathcal{L}\left(\mathbf{a},\mathbf{\Theta}\right), \quad \quad \text{s.t.}\ \|\mathbf{a}\|_0 \leq \kappa
\label{eq:gaussianspa_objective}
\end{equation}
where $\mathbf{\Theta}$ denotes all primitive parameters except opacity $\mathbf{a}$, and $\kappa$ is the primitive budget.
The optimization alternates between a standard reconstruction step and a sparsification step that gradually reduces the opacity of expendable primitives.
GaussianSpa applies this scheme during training; we repurpose it as a post-training refinement stage of $5$k iterations.
This allows the remaining primitives to compensate for the reduced primitive set by redistributing opacity and geometry among nearby Gaussians.
In this stage, we additionally apply the effective rank regularization term $\mathcal{L}_{\text{erank}}$ defined in \cref{eq:erank_reg}.

\paragraph{Second Pruning and Fine-Tuning.}
For a second pruning pass, we recompute $\Delta SE_k$ to remove primitives that have become redundant. Finally, another $5$k iterations of fine-tuning recover quality lost in the previous steps, for a total of $10$k post-processing iterations.

\subsection{Encoding with \ourCompression}
\label{sec:encoding}

Following compaction with \ourCompaction, the remaining primitives are encoded by \ourCompression, an image-based codec extending the Self-Organizing Gaussians (SOG) approach~\cite{morgenstern2024compact} with two novel contributions described below.

The core idea of SOG~\cite{morgenstern2024compact} is to arrange the $N$ primitives into a $\lceil\sqrt{N}\rceil \times \lceil\sqrt{N}\rceil$ 2D grid using PLAS, a parallel assignment algorithm that preserves spatial locality in the grid layout.
The resulting attribute maps are locally smooth, making them well-suited for standard image codecs.
SOG stores the individual per-attribute slices as lossless JPEG-XL images, with quantization applied post-training.
In the original SOG method, densification parameters, training-time smoothness regularization, and quantization are coupled, which makes its encoding stage difficult to apply in isolation.
The \texttt{.sog} format keeps the image-based storage idea of SOG~\cite{morgenstern2024compact}, but makes it usable as a post-training format by using Morton-order curves to arrange primitives and k-means clustering for spherical harmonics dimensionality reduction, with centroids sorted lexicographically.

Like \texttt{.sog}, SOG-XT decouples encoding from reconstruction and compaction entirely, operating as a pure post-training step on any compacted scene.
SOG-XT retains the PLAS-based grid layout for all attributes but replaces the 48 raw SH slices of SOG~\cite{morgenstern2024compact} with vector quantization: k-means clustering reduces the high-dimensional SH representation to a compact codebook, with only per-primitive cluster indices stored.
Rather than sorting these centroids lexicographically as in \texttt{.sog}, SOG-XT sorts the 2D cluster centroids with PLAS, yielding a codebook that is itself locally smooth in 2D.
Cluster assignments are stored as two 8-bit image channels $(u, v)$ indexing the PLAS-sorted centroid grid.
This deliberately caps the SH codebook at $256 \times 256 = \num{65536}$ entries, which was the point where increasing the codebook size no longer improved PSNR in our experiments.
This 2D structure additionally enables fine-tuning to refine the codebook and provides a natural foundation for future vector quantization methods that exploit spatial coherence more aggressively.

\subsection{Covariance Symmetry and \covopt}

The mapping from a 3D covariance matrix to the parameterization of tuples of quaternions and scales $(q, s)$ is one-to-many.
A unique covariance matrix maps to an entire equivalence class of discrete parameter choices that all yield the identical spatial distribution.
This symmetry stems from three independent sources of redundancy in the decomposition of $\mathbf\Sigma$: \textbf{Scale Permutations}:
The three principal radii can be assigned to the coordinate axes in $3!=6$ ways, provided they are compensated by $90^\circ$ rotations around principal axes to arrive at the identical covariance.
\textbf{Eigenvector Orientations}: Flipping eigenvector directions yields $2^3$ combinations, of which 4 represent valid rotations satisfying $\det(R)=1$.
\textbf{Quaternion Double-Cover}: For every valid rotation matrix $R$, there exist two quaternions, $q$ and $-q$, which produce that exact rotation.

Combined, this results in $6 \cdot4 \cdot 2 = 48$ distinct parameter combinations within an equivalence class that describe a single Gaussian.
To exploit this symmetry, we propose \textbf{\covoptLong (\covopt)}. 
Similarly to \textit{PLAS}, \textit{\covopt} employs coarse to fine optimization over 2D parameter grids. First, to ensure comparable magnitudes during optimization, the floating-point quaternions and log-scales are mapped to 8-bit quantized proxy grids. 
The algorithm then iterates over a sequence of exponentially decreasing blur radii. 
At each scale, a low-pass filter is applied to the current proxies to generate spatially smoothed target grids. For every primitive, \textit{\covopt} evaluates all candidates within its equivalence class and assigns the representative that minimizes the sum of absolute channel differences, i.e. the $\ell_1$ distance, to the smoothed targets. 
Because the search space is strictly restricted to valid equivalence classes, \textit{\covopt} effectively removes artificial variance to improve the compressibility of scale and quaternion grids while leaving the resulting 3D covariances and grid coordinates unchanged. 

\subsection{\ourCompression format}
\begin{figure}[tb]
  \centering
  \includegraphics[width=\linewidth]{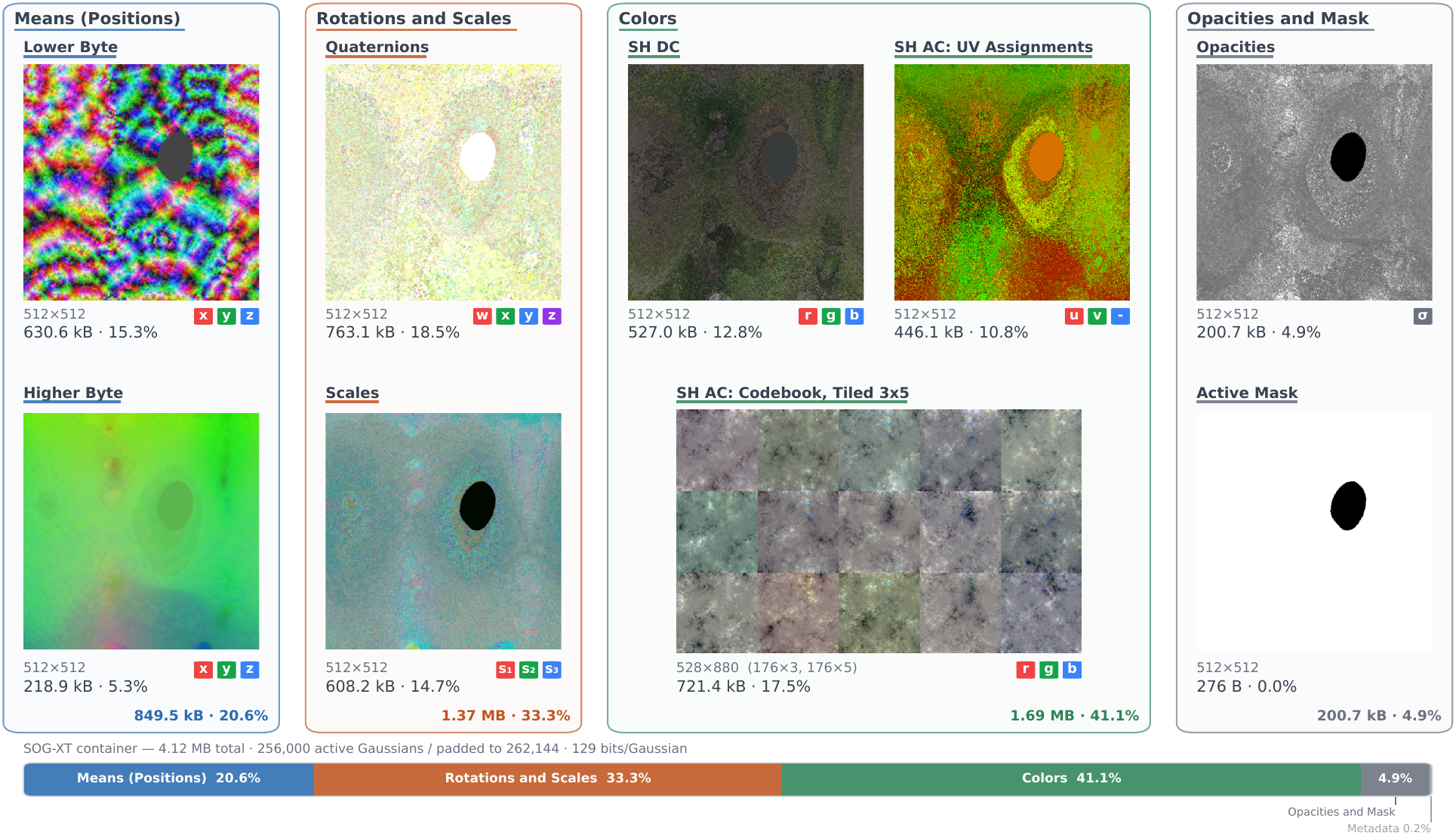}
  \caption{Encoding overview for \ourCompression. PLAS sorts primitives into 2D attribute grids using both mean byte planes. View dependent color is reduced with a k-means codebook whose centroids are PLAS sorted again, and assignments are stored as UV indices. Bars show the relative on-disk size contribution of each component.}
  \label{fig:compression_grid}
\end{figure}
~\Cref{fig:compression_grid} shows an example of an encoded \ourCompression container with the file size and share of the total for each stored component.

Following SOG~\cite{morgenstern2024compact},  we arrange primitives into a 2D grid with PLAS and store the attribute grids as images.
SOG uses JPEG-XL with higher bit depth support, while we restrict storage to browser-decodable 8-bit images.
We avoid truncation by padding primitives to a square grid with side length aligned to a multiple of 8 and storing an \emph{active mask}.
Padded primitives use fixed defaults and are dropped at decoding time, which adds little metadata overhead.

We apply per-channel min-max quantization with uniform step sizes and store each channel's minimum and maximum as float32 values in a YAML metadata file.
In the example container shown in \cref{fig:compression_grid}, all metadata including these ranges accounts for only $0.2\%$ of the total size on disk.
These values are therefore negligible for compression, but necessary for exact dequantization.

For the Gaussian means, which define primitive positions, we follow SOG and apply a signed log remapping for space contraction before 16-bit quantization.
As in browser-oriented \texttt{.sog}, we store the quantized positions as two 8-bit byte planes.
The upper byte captures coarse spatial structure, while the lower byte contains finer residual variation and is less smooth under image coding.
Our contribution is to expose this split to the layout optimization: we run PLAS on both byte planes, with a stronger weight on the upper byte and a weaker weight on the lower byte, and apply the resulting permutation to all remaining attributes.

For view-dependent color, we split spherical harmonics into the DC term $f_\mathrm{dc}$ and the remaining coefficients $f_\mathrm{rest}$.
We store $f_\mathrm{dc}$ directly as an 8-bit image.
We compress $f_\mathrm{rest}$ with a k-means codebook of size $K$ that scales with the number of active primitives.
In our experiments, we choose the SH codebook size as $K=S^2$.
We set the codebook side length $S$ from $\sqrt{N/8}$, clamp it to $[8,256]$, and round it down to a multiple of $8$.
As a further enhancement, we PLAS-sort the SH codebook centroids to obtain a 2D centroid grid and store assignments as two 8-bit UV channels that index this grid.
We quantize the SH codebook centroids with 100 levels using per-channel min-max ranges.
Each centroid contains 15 non-DC SH coefficients with RGB values.
We store the resulting centroid grid as a single RGB image by arranging the 15 coefficient images in a $3\times5$ grid of tiles.

Scales use a log mapping and 8-bit quantization, and opacities use an inverse sigmoid mapping and 8-bit quantization.
We optimize quaternions jointly with scales using \covopt{}, which enables quaternion quantization with $q=99$.
Without \covopt{}, we find that $q=255$ is needed for comparable quality.
The encoder does not require camera parameters or a renderer.

\subsection{Adapting the \ourCompression Format in Encoding-aware Fine-tuning}
\label{sec:finetuning}

Optionally, we fine-tune with \ourCompression in the loop.
At each training step, we map the current Gaussian attributes to the quantized image tensors used by \ourCompression, reconstruct a Gaussian set from these tensors, render it, and backpropagate the rendering loss through this round trip.
The discrete codec structure is fixed at the start, including the active mask, PLAS order, SH assignments, and codebook layout.
During training, SH assignments stay fixed while centroid values are recomputed from the current $f_\mathrm{rest}$ coefficients.
Rounding is handled with a straight-through estimator~\cite{bengio2013estimating}.

We optimize the standard rendering objective and add lightweight smoothness losses on the quantized images.
We also re-run \covopt{} during fine-tuning to preserve the compressibility benefits of covariance symmetry resolution as the parameters change.
The output format is identical to \ourCompression without fine-tuning, so the decoder remains unchanged.
Implementation details are provided in the appendix.

\section{Evaluation}
\label{sec:eval}

Let us demonstrate the effectiveness of keeping things simple: we will show how \ourCompaction improves reconstruction quality with a reduced number of primitives compared to INRIA 3DGS and competing compaction methods. Then we'll look into our individual contributions that make up \ourCompression: 2D sorted codebooks and our PRAS covariance smoothing. Finally, we show how our results compare to the previous best-in-class, HAC++\cite{chen2025hac-plus}, over the standard datasets for benchmarking 3DGS compression.

\subsection{Evaluation protocol for 3DGS compression}
The original 3DGS paper~\cite{kerbl3Dgaussians} established a de-facto evaluation protocol through its public codebase, which is the baseline for many follow-up works.
As explicitly stated in the 3DGS.zip survey~\cite{3DGSzip2025}, this protocol specifies the datasets and splits (21 scenes across \TandT, \MipNeRF, \DeepBlending, and \SyntheticNeRF), scene-dependent training resolutions ($4\times$ downsampling for \MipNeRF outdoor scenes, $2\times$ for indoor), a fixed evaluation resolution (images downscaled to 1600 px on the longest side), and evaluation with peak signal-to-noise ratio (PSNR), structural similarity index measure (SSIM)~\cite{wang2004image}, and learned perceptual image patch similarity (LPIPS)~\cite{zhang2018perceptual}, using the VGG backbone for LPIPS and images in their $[0,1]$ range.
Full details are provided in the appendix.

Deviating from this evaluation protocol can lead to considerable differences in reported metrics.
The effects of training and test resolutions are discussed in 3DGS.zip survey\cite{3DGSzip2025}, while a discussion of the incorrect LPIPS range being used in 3DGS literature can be found in the NeRF baselines benchmark\cite{kulhanek2025nerfbaselines}.

In all tables that have color-highlighted values, the best results in each category are highlighted \colorbox{lightred}{first}, \colorbox{lightorange}{second}, and \colorbox{lightyellow}{third}. 

\subsection{Baselines: INRIA for Quality and HAC++ for Compression}

For an uncompressed quality reference, we use INRIA 3DGS~\cite{kerbl3Dgaussians}, which we denote INRIA-Q.
The original 3DGS protocol trains for 30k iterations, which matches the reconstruction stage of our pipeline.
Our compaction stage then adds 10k refinement iterations before compression.
We therefore train the INRIA quality reference for 40k iterations as well, matching the total pre-compression optimization budget of our method.

As the compression baseline, we use HAC++~\cite{chen2025hac-plus}, the highest ranked method in the 3DGS.zip survey.
Recomputing all published 3DGS compression methods under a single protocol is infeasible, but HAC++ is the strongest available reference point in the survey.
Only the preprint HEMGS~\cite{liu2024hemgs} achieves higher PSNR on several datasets, while HAC++ is more balanced across metrics and has public code available.
We therefore recompute HAC++ under our evaluation protocol and compare to the remaining methods using their self-reported values from the survey.
We run HAC++ for 44k iterations as a compute-matched baseline.
This matches the total budget of the full KISS-GS pipeline, whose stages use 30k iterations for reconstruction, 10k iterations for POPSpa compaction, and 4k iterations for optional encoding-aware fine-tuning.
We use the parameters provided by the authors for their high- and low-rate versions, but follow our evaluation protocol throughout, which leads to slight differences from the numbers reported in their paper.
The appendix ablates HAC++ across the reported 30k setting and our 30k, 40k, and 44k recomputations under both evaluation protocols.

\subsection{Evaluating Compaction with \ourCompaction}

~\Cref{fig:main_rd} shows results for \ourCompaction using seven Gaussian primitive budgets.
For the smallest model, we set the target to \num{64}\,\text{k} Gaussians for all datasets, except for \SyntheticNeRF, where the lower scene complexity motivates a target of \num{16}\,\text{k}.
To obtain an intended pruning ratio of approximately $80\%$, we first train an over-parameterized 3DGS model with \emph{gsplat-MCMC}~\cite{ye2025gsplat} using \num{320}\,\text{k} primitives (\num{80}\,\text{k} for \SyntheticNeRF).

For each subsequent point in ~\cref{fig:main_rd}, we double both the compacted and the corresponding initial budget.
The initial (uncompacted) models are capped by a dataset-specific maximum of \num{3000}\,\text{k} Gaussians for \TandT and \MipNeRF, \num{2048}\,\text{k} for \DeepBlending, and \num{640}\,\text{k} for \SyntheticNeRF.
The right-most point corresponds to the maximal model size without pruning and only applies \num{10}\,\text{k} fine-tuning iterations.

\begin{table*}[!htb]

\caption{Comparison of 3DGS compaction methods across four standard benchmark datasets. Ranks represent average rankings across all available datasets, with model size and overall quality weighted equally: size contributes half the weight, and PSNR, SSIM, and LPIPS each contribute one-sixth. 
In the top half, post-training pruning is applied to a 30k-step vanilla 3DGS checkpoint, removing 90\% of Gaussians. The bottom half compares against compaction-aware training methods, where each method is trained for 10k steps beyond its original implementation, resulting in 40k steps for all methods except Mini-Splatting 2 which was originally trained for only 18k steps. Our method is post-training only, applied to a checkpoint trained with gsplat's MCMC implementation.}

\begin{adjustbox}{width=12.2cm,center}

\newcommand{\gc}{\cellcolor{gray!25}}

\begin{tabular}{c ll|lllr|lllr|lllr|lllr}
& Method & Rank & \multicolumn{4}{c|}{Tanks and Temples} & \multicolumn{4}{c|}{Mip-NeRF 360} & \multicolumn{4}{c|}{Deep Blending} & \multicolumn{4}{c}{Synthetic NeRF} \\
&  & \tiny  & \tiny PSNR$\uparrow$ & \tiny SSIM$\uparrow$ & \tiny LPIPS$\downarrow$ & \tiny k Gauss & \tiny PSNR$\uparrow$ & \tiny SSIM$\uparrow$ & \tiny LPIPS$\downarrow$ & \tiny k Gauss & \tiny PSNR$\uparrow$ & \tiny SSIM$\uparrow$ & \tiny LPIPS$\downarrow$ & \tiny k Gauss & \tiny PSNR$\uparrow$ & \tiny SSIM$\uparrow$ & \tiny LPIPS$\downarrow$ & \tiny k Gauss \\
\bottomrule

\multirow{4}{*}[-0.5em]{\begin{sideways}\shortstack{post\\training}\end{sideways}}
& Inria~\cite{kerbl3Dgaussians}~\venuetag{SIGGRAPH~'23} (30k) &   & 23.77 & .847 & .170 & 1,563 & 27.28 & .807 & .222 & 2,723 & 29.69 & .905 & .242 & 2,463 & 33.46 & .970 & .030 & 259 \\
\cmidrule(lr){2-19}

& \gc + Ours (+10k) & \cellcolor{lightred}1.1 & \cellcolor{lightorange}23.37 & \cellcolor{lightred}.819 & \cellcolor{lightred}.231 & \cellcolor{lightred}156 & \cellcolor{lightorange}26.78 & \cellcolor{lightred}.781 & \cellcolor{lightred}.289 & \cellcolor{lightred}272 & \cellcolor{lightred}29.51 & \cellcolor{lightred}.903 & \cellcolor{lightorange}.265 & \cellcolor{lightred}246 & \cellcolor{lightred}32.46 & \cellcolor{lightred}.964 & \cellcolor{lightred}.041 & \cellcolor{lightred}25 \\

& + Speedy-Splat~\cite{HansonSpeedy}~\venuetag{CVPR~'25} (+10k)  & \cellcolor{lightorange}1.5 & \cellcolor{lightred}23.52 & \cellcolor{lightorange}.813 & \cellcolor{lightorange}.245 & \cellcolor{lightred}156 & \cellcolor{lightred}26.87 & \cellcolor{lightorange}.772 & \cellcolor{lightyellow}.305 & \cellcolor{lightred}272 & \cellcolor{lightorange}29.45 & \cellcolor{lightorange}.902 & \cellcolor{lightyellow}.267 & \cellcolor{lightred}246 & \cellcolor{lightorange}32.40 & \cellcolor{lightorange}.962 & \cellcolor{lightorange}.045 & \cellcolor{lightred}25 \\

& \gc + PUP 3D-GS~\cite{HansonTuPUP3DGS}~\venuetag{CVPR~'25} (+10k)  & \cellcolor{lightyellow}1.8 & \cellcolor{lightyellow}22.62 & \cellcolor{lightyellow}.798 & \cellcolor{lightyellow}.251 & \cellcolor{lightred}156 & \cellcolor{lightyellow}25.58 & \cellcolor{lightyellow}.767 & \cellcolor{lightorange}.300 & \cellcolor{lightred}272 & \cellcolor{lightyellow}29.26 & \cellcolor{lightred}.903 & \cellcolor{lightred}.264 & \cellcolor{lightred}246 & \cellcolor{lightyellow}31.49 & \cellcolor{lightyellow}.958 & \cellcolor{lightyellow}.049 & \cellcolor{lightred}25 \\

\midrule

\multirow{4}{*}[-1em]{\begin{sideways}\shortstack{compaction\\aware}\end{sideways}}
& gsplat-MCMC~\cite{ye2025gsplat} (30k) &   & 24.54 & .869 & .150 & 1,280 & 27.95 & .829 & .196 & 2,560 & 30.08 & .907 & .244 & 1,280 & 33.82 & .972 & .028 & 320 \\

& \gc + Ours (+10k) & \cellcolor{lightred}1.7 & \cellcolor{lightred}24.47 & \cellcolor{lightred}.856 & \cellcolor{lightyellow}.179 & \cellcolor{lightred}256 & \cellcolor{lightred}27.79 & \cellcolor{lightyellow}.811 & \cellcolor{lightyellow}.239 & \cellcolor{lightred}512 & \cellcolor{lightyellow}29.94 & \cellcolor{lightyellow}.904 & \cellcolor{lightyellow}.262 & \cellcolor{lightred}256 & \cellcolor{lightred}33.41 & \cellcolor{lightred}.969 & \cellcolor{lightorange}.032 & \cellcolor{lightorange}64 \\
\cmidrule(lr){2-19}

& gsplat-MCMC~\cite{ye2025gsplat} (40k) & \cellcolor{lightorange}2.5 & \cellcolor{lightyellow}23.90 & .841 & .205 & \cellcolor{lightred}256 & \cellcolor{lightorange}27.52 & .809 & .243 & \cellcolor{lightred}512 & 29.43 & .899 & .270 & \cellcolor{lightred}256 & \cellcolor{lightyellow}32.73 & \cellcolor{lightyellow}.967 & \cellcolor{lightyellow}.036 & \cellcolor{lightorange}64 \\

& \gc GaussianSpa~\cite{zhang2025gaussianspa}~\venuetag{CVPR~'25} (40k) & \cellcolor{lightyellow}2.8 & 23.63 & \cellcolor{lightorange}.849 & \cellcolor{lightred}.171 & \cellcolor{lightyellow}349 & \cellcolor{lightyellow}27.40 & \cellcolor{lightred}.817 & \cellcolor{lightorange}.227 & \cellcolor{lightorange}561 & \cellcolor{lightred}30.09 & \cellcolor{lightred}.913 & \cellcolor{lightorange}.245 & \cellcolor{lightyellow}457 & \cellcolor{lightorange}33.34 & \cellcolor{lightorange}.968 & \cellcolor{lightred}.029 & \cellcolor{lightyellow}212 \\

& Mini-Splatting2~\cite{fang2024minisplattingv2} (28k) & 3.3 & 23.57 & \cellcolor{lightyellow}.843 & \cellcolor{lightorange}.178 & 357 & 27.33 & \cellcolor{lightorange}.815 & \cellcolor{lightred}.223 & \cellcolor{lightyellow}619 & \cellcolor{lightorange}30.05 & \cellcolor{lightorange}.911 & \cellcolor{lightred}.241 & 658 & 32.69 & .964 & .037 & \cellcolor{lightred}54 \\

& \gc Taming 3DGS~\cite{taming20243dgs}~\venuetag{SIGGRAPH~Asia~'24} (40k) & \gc 3.9 & \cellcolor{lightorange}24.10 & \gc .836 & \gc .203 & \cellcolor{lightorange}318 & \gc 27.33 & \gc .795 & \gc .258 & \gc 684 & \gc 29.77 & \gc .903 & \gc .273 & \cellcolor{lightorange}294 & \gc  & \gc  & \gc  & \gc  \\

\end{tabular}
\end{adjustbox}

\label{tab:compaction}
\end{table*}

\Cref{tab:compaction} presents results for both post-training pruning and compaction-aware training. 
In the post-training setting, our method consistently outperforms all competing approaches across all benchmarks. 

The compaction-aware comparison is more nuanced: since most methods cannot easily predetermine the final scene size or quality, the compared samples sit at slightly different positions on each method's quality-size trade-off curve. 
Despite this, our method achieves the best average rank of 1.7 while operating at significantly fewer Gaussians than competing methods across most datasets. 
The exception is \SyntheticNeRF, where model sizes are more comparable. 
In \DeepBlending, we achieve competitive quality using only 56\% of the primitives of the best performing method. 
Full quality-size trade-off curves for all datasets are provided in the appendix.
Across these curves, our method consistently achieves higher quality than competing methods at a given size.

\subsection{Evaluating Compression with \ourCompression}
As shown in \cref{fig:main_rd} for \MipNeRF and in the appendix for other datasets, our encoding method \ourCompression is able to compress the test scenes to single-digit Megabyte sizes. 
With a single parameter choice, as presented in \cref{sec:encoding}, our method works well across a large range of primitive counts, from several thousand to millions. The same encoding method works on the INRIA 40k baseline, highlighting the modularity of the approach. 
Two popular post-training encoding methods in \texttt{.spz} and \texttt{.sog} are benchmarked against our method using their official implementations (\texttt{splat-transform} v1.6.0~\cite{playcanvas_sog} for \texttt{.sog}, the official \texttt{spz} PyPI package v0.0.6~\cite{niantic2024spz} for \texttt{.spz}). 
With \texttt{.sog}, the files are even larger, and this format appears sensitive to the number of primitives, producing a jagged rate-distortion curve.

\begin{table*}[!htb]

\caption{Ablation of contributions to the \ourCompression format on the Bicycle scene with \num{256}\,\text{k} primitives.
The first row contains the absolute values after encoding. The following rows disable individual features of \ourCompression, ordered by size penalty. Most removals increase the final size considerably at small quality gains. Using image encoding, our novel codebook sorting and its UV label packing bring clean size wins without incurring any quality loss. Find additional rows in the appendix.}

\begin{adjustbox}{width=10.5cm,center}
\rowcolors{1}{gray!25}{white}

\small
\setlength{\tabcolsep}{4pt}
\small
\setlength{\tabcolsep}{4pt}
\begin{tabular}{p{5.5cm} r r r r r}
\hiderowcolors %
& & \multicolumn{4}{c}{\textbf{Effect of Leaving Feature Out}} \\
\cmidrule(lr){3-6}
& Size [MB] & \makecell[r]{$\Delta$ Size [Bytes] $\downarrow$} & \makecell[r]{$\Delta$ PSNR $\uparrow$} & \makecell[r]{$\Delta$ SSIM $\uparrow$} & \makecell[r]{$\Delta$ LPIPS $\downarrow$} \\
\cmidrule(lr){2-6}
\textbf{SOG-XT (Full Pipeline)} & 3.878 & (3,877,944) & (24.2281) & (0.68176) & (0.35404) \\
\midrule
\showrowcolors %
\noalign{\global\rownum=1} %
w/o SH AC codebooks & 7.313 & \textcolor{DiffRed}{+3,434,816} & \textcolor{DiffGreen}{+0.2320} & \textcolor{DiffGreen}{+0.00790} & \textcolor{DiffGreen}{-0.00579} \\
w/o WebP image compression & 6.908 & \textcolor{DiffRed}{+3,030,148} & \textcolor{DiffGray}{+0.0000} & \textcolor{DiffGray}{+0.00000} & \textcolor{DiffGray}{+0.00000} \\
w/o quaternion u8 quantization & 6.427 & \textcolor{DiffRed}{+2,548,718} & \textcolor{DiffGreen}{+0.0317} & \textcolor{DiffGreen}{+0.00147} & \textcolor{DiffGreen}{-0.00067} \\
w/o primitive PLAS sorting & 4.493 & \textcolor{DiffRed}{+614,851} & \textcolor{DiffGreen}{+0.0059} & \textcolor{DiffGreen}{+0.00022} & \textcolor{DiffGreen}{-0.00036} \\
w/o centroid PLAS sorting & 3.920 & \textcolor{DiffRed}{+42,416} & \textcolor{DiffGray}{+0.0000} & \textcolor{DiffGray}{+0.00000} & \textcolor{DiffGray}{+0.00000} \\
w/o label UV packing & 3.890 & \textcolor{DiffRed}{+12,142} & \textcolor{DiffGray}{+0.0000} & \textcolor{DiffGray}{+0.00000} & \textcolor{DiffGray}{+0.00000} \\
w/o \covopt & 3.866 & \textcolor{DiffGreen}{-11,810} & \textcolor{DiffRed}{-0.1104} & \textcolor{DiffRed}{-0.00538} & \textcolor{DiffRed}{+0.00273} \\
w/o means signed-log remapping & 3.359 & \textcolor{DiffGreen}{-519,272} & \textcolor{DiffRed}{-7.5508} & \textcolor{DiffRed}{-0.37864} & \textcolor{DiffRed}{+0.22151} \\
\end{tabular}
\end{adjustbox}

\label{tab:compression-ablation}

\end{table*}

In \Cref{tab:compression-ablation}, we show how the individual components of \ourCompression contribute to its high compression ratio.
In these leave-one-out experiments, we can see that most components provide size wins against small quality losses, like the quantization operations.
Our 2D codebooks (centroids PLAS sorting and label UV packing) bring free size wins without any quality loss.
Our \covopt method incurs a small size hit, but provides large quality gains.
The highest quality loss occurs from building codebooks for SH AC, but otherwise, the files would nearly triple in size.
Without the log-space contraction of means, file sizes would shrink.
However, with 16-bit quantization for the means, reconstruction quality would degrade severely.

\subsection{Adaptation and Comparison with State of the Art}

In \Cref{fig:main_rd}, we show that encoding-aware fine-tuning clearly improves the rate-distortion curve over post-training compression, achieving another $2.2\times$ reduction at INRIA quality for the \MipNeRF dataset.
While \ourCompression by itself can already achieve great results without feedback from the renderer, running some adaptation can be worth it if a GPU and training views are available at encoding time.
A sensitivity analysis in the appendix shows that the default fine-tuning setting is not a fragile optimum; the SH codebook size is the clearest rate-control knob, while additional fine-tuning steps and stronger TV regularization do not improve the trade-off.

\begin{table*}[!htb]

\caption{File-size reduction factors over INRIA 3DGS after 40k iterations.
Each entry is the interpolated point where a method matches INRIA-Q for the corresponding metric.
KISS-GS achieves the highest reductions on \TandT and \MipNeRF, and on two of the three \SyntheticNeRF metrics.}

\begin{adjustbox}{width=12.2cm,center}
\rowcolors{2}{gray!25}{white}

\colorlet{selfreportgray}{black!80}

\begin{tabular}{l |ccc|ccc|ccc|ccc}
\toprule
Method & \multicolumn{3}{c|}{Tanks and Temples} & \multicolumn{3}{c|}{Mip-NeRF 360} & \multicolumn{3}{c|}{Deep Blending} & \multicolumn{3}{c}{Synthetic NeRF} \\
 & \tiny PSNR & \tiny SSIM & \tiny LPIPS & \tiny PSNR & \tiny SSIM & \tiny LPIPS & \tiny PSNR & \tiny SSIM & \tiny LPIPS & \tiny PSNR & \tiny SSIM & \tiny LPIPS \\
\midrule
\textcolor{selfreportgray}{CodecGS~\cite{lee2025compression3dgaussiansplatting}~\venuetag{ICCV '25}\textsuperscript{$\dagger$}} & \textcolor{selfreportgray}{53*} & \textcolor{selfreportgray}{-} & \textcolor{selfreportgray}{-} & \cellcolor{lightyellow}\strut \textcolor{selfreportgray}{72*} & \cellcolor{lightyellow}\strut \textcolor{selfreportgray}{72*} & \textcolor{selfreportgray}{-} & \textcolor{selfreportgray}{76*} & \textcolor{selfreportgray}{76*} & \textcolor{selfreportgray}{-} & \textcolor{selfreportgray}{n/a} & \textcolor{selfreportgray}{n/a} & \textcolor{selfreportgray}{n/a} \\
\textcolor{selfreportgray}{ContextGS~\cite{wang2024contextgs}~\venuetag{NeurIPS '24}\textsuperscript{$\dagger$}} & \textcolor{selfreportgray}{42*} & \textcolor{selfreportgray}{42*} & \textcolor{selfreportgray}{-} & \textcolor{selfreportgray}{55*} & \textcolor{selfreportgray}{55*} & \textcolor{selfreportgray}{-} & \cellcolor{lightorange}\strut \textcolor{selfreportgray}{187*} & \cellcolor{lightorange}\strut \textcolor{selfreportgray}{187*} & \textcolor{selfreportgray}{-} & \textcolor{selfreportgray}{n/a} & \textcolor{selfreportgray}{n/a} & \textcolor{selfreportgray}{n/a} \\
\textcolor{selfreportgray}{gsplat-1M~\cite{hu2024gsplat}~\venuetag{arXiv '24}\textsuperscript{$\dagger$}} & \textcolor{selfreportgray}{60*} & \textcolor{selfreportgray}{44} & \cellcolor{lightyellow}\strut \textcolor{selfreportgray}{26} & \textcolor{selfreportgray}{52} & \textcolor{selfreportgray}{57} & \cellcolor{lightyellow}\strut \textcolor{selfreportgray}{28} & \textcolor{selfreportgray}{n/a} & \textcolor{selfreportgray}{n/a} & \textcolor{selfreportgray}{n/a} & \textcolor{selfreportgray}{n/a} & \textcolor{selfreportgray}{n/a} & \textcolor{selfreportgray}{n/a} \\
\textcolor{selfreportgray}{HAC~\cite{chen2024hac}~\venuetag{ECCV '24}\textsuperscript{$\dagger$}} & \textcolor{selfreportgray}{49*} & \textcolor{selfreportgray}{48} & \textcolor{selfreportgray}{-} & \textcolor{selfreportgray}{46*} & \textcolor{selfreportgray}{46*} & \textcolor{selfreportgray}{-} & \textcolor{selfreportgray}{150*} & \textcolor{selfreportgray}{129} & \textcolor{selfreportgray}{-} & \cellcolor{lightorange}\strut \textcolor{selfreportgray}{47} & \textcolor{selfreportgray}{-} & \textcolor{selfreportgray}{-} \\
\textcolor{selfreportgray}{HEMGS~\cite{liu2024hemgs}~\venuetag{arXiv '24}\textsuperscript{$\dagger$}} & \textcolor{selfreportgray}{69*} & \cellcolor{lightyellow}\strut \textcolor{selfreportgray}{69*} & \textcolor{selfreportgray}{-} & \textcolor{selfreportgray}{59*} & \textcolor{selfreportgray}{59*} & \textcolor{selfreportgray}{-} & \cellcolor{lightred}\strut \textcolor{selfreportgray}{229*} & \cellcolor{lightred}\strut \textcolor{selfreportgray}{229*} & \textcolor{selfreportgray}{-} & \cellcolor{lightred}\strut \textcolor{selfreportgray}{57*} & \textcolor{selfreportgray}{-} & \textcolor{selfreportgray}{-} \\
HAC++~\cite{chen2025hac-plus}~\venuetag{TPAMI '25} & \cellcolor{lightyellow}\strut 77* & 68 & - & 70 & 36 & - & \cellcolor{lightyellow}\strut 156 & \cellcolor{lightyellow}\strut 156 & - & - & - & - \\
Ours (SOG-XT) & \cellcolor{lightorange}\strut 227 & \cellcolor{lightorange}\strut 104 & \cellcolor{lightorange}\strut 60 & \cellcolor{lightorange}\strut 104 & \cellcolor{lightorange}\strut 73 & \cellcolor{lightorange}\strut 40 & - & - & - & 10 & - & - \\
Ours+FT & \cellcolor{lightred}\strut 319 & \cellcolor{lightred}\strut 131 & \cellcolor{lightred}\strut 72 & \cellcolor{lightred}\strut 228 & \cellcolor{lightred}\strut 109 & \cellcolor{lightred}\strut 64 & 85 & 98 & - & \cellcolor{lightyellow}\strut 21 & \cellcolor{lightred}\strut 14 & \cellcolor{lightred}\strut 14 \\
\bottomrule
\end{tabular}

\end{adjustbox}

\newcommand{\tabnote}[2]{%
  \par\noindent\footnotesize
  \hangindent=2.2em\hangafter=1%
  \makebox[2.2em][l]{#1}#2\par
}
\begin{adjustwidth}{1.2em}{0em} %
\footnotesize
\tabnote{*}{smallest available point already reaches INRIA-Q, so the value is a lower bound.}
\tabnote{-}{INRIA-Q not reached within available points.}
\tabnote{\textsuperscript{$\dagger$}}{self-reported result from 3DGS.zip, not recomputed under our protocol.}
\end{adjustwidth}

\label{tab:reduction-factors}

\end{table*}

In \Cref{tab:reduction-factors}, each entry is the interpolated file-size reduction at which a method matches the INRIA 3DGS 40k quality reference.
We use recomputed HAC++ as the main baseline, since it is the strongest reproducible published method in the 3DGS.zip survey.
HEMGS reports higher reductions in some cases, but it is a preprint without public code, so we include it only as self-reported context.
On \TandT and \MipNeRF, KISS-GS reaches INRIA-Q at substantially higher reduction factors than HAC++ across all reported metrics.
On \DeepBlending, the gap is already visible before SOG-XT is applied: compressed HAC++ exceeds uncompressed gsplat-MCMC and INRIA 3DGS on \textit{Playroom} (30.63 vs.\ 30.42 and 30.09) and \textit{Dr.~Johnson} (29.60 vs.\ 29.48 and 29.43).
This points to HAC++'s anchor-based learned representation and regularization rather than to SOG-XT quantization or POPSpa pruning.
To keep the comparison compact, we include literature methods only if they provide at least four finite entries in this table.

Notably, many competing methods from the literature do not reach INRIA quality in the perceptual LPIPS metric at any of their available operating points.
This may point to overfitting the PSNR metric for the training views and losing generalization capabilities in rendering novel views or novel resolutions (for the \MipNeRF dataset).

Runtime measurements in the appendix show that post-hoc \ourCompression encoding is inexpensive compared with reconstruction and compaction.
Encoding-aware fine-tuning is more expensive, but it can be omitted.
A reference Python decoder reconstructs models with up to 1024k Gaussians in well under a second on CPU, supporting our claim that deployment remains simple.

\section{Have we kept it simple?}
\label{sec:conclusion}

The title of this paper, together with the Dijkstra quote in the introduction, makes a deliberately cheeky promise.
We set out to keep 3DGS compression simple.
With respect to simplicity, we fell short in one place: the encoder is a sophisticated modular pipeline with modules such as \ourCompaction and novel components such as \covopt.
This deliberate complexity lets KISS-GS separate reconstruction, compaction, encoding, and \textbf{encoding-aware fine-tuning}, close the \textbf{attribution gap}, and make the source of each gain measurable.

The payoff is substantial: KISS-GS reaches \textbf{single-digit megabytes} at vanilla 3DGS reconstruction quality.
With $319\times$ size reduction on \TandT in PSNR and $228\times$ on \MipNeRF, it sets a new benchmark for 3DGS compression on these two real-world datasets.
A pipeline built around a \textbf{simple image-based encoding format} can exceed the strongest published baselines on the main real-world benchmarks while keeping deployment simple.
\ourCompression reconstructs standard 3D Gaussian Splatting attributes through standard image decoding and small, deterministic rescaling and reshaping operations.
This keeps implementation effort low and makes decoders easy to port across languages and platforms.
Optional \textbf{encoding-aware fine-tuning} improves rate-distortion without changing this decoder.
\DeepBlending remains the main boundary, where anchor-based learned representations better absorb capture inconsistencies than the standard-splat reconstructions used as input to KISS-GS.
Its modular design makes future advances in reconstruction, compaction, and vector quantization directly useful for compression.
Still, KISS-GS shows that state-of-the-art 3DGS compression does not require a complex deployment stack: the strongest results can come from a sophisticated encoder while keeping the decoder deliberately \textit{simple}.

\section*{Acknowledgements}
This work has partly been funded by the European Union’s Horizon Europe research and innovation program (Luminous, grant no. 101135724), the German Research Foundation (3DIL, grant no. 502864329), and the Fraunhofer society (Self Structured Gaussians).

\clearpage %
\bibliographystyle{splncs04}
\bibliography{main}

\clearpage
\setcounter{page}{1}

\section{Appendix to KISS-GS: 3D Gaussian Splatting Compression Kept Simple}
\label{sec:appendix-rationale}

In the main paper, we have shown how to build a modular pipeline for 3DGS compression whose output is a compact \ourCompression container that remains simple to decode while still reaching record-setting compression levels.
With the additional material in this appendix, we strengthen the paper's claims with more detailed comparisons across datasets and metrics, additional ablations, a visual explanation of the covariance symmetry, and qualitative comparisons of rendered views.

\subsection{Runtime Performance}

\begin{figure}[htb]
  \centering
  \includegraphics[width=\linewidth]{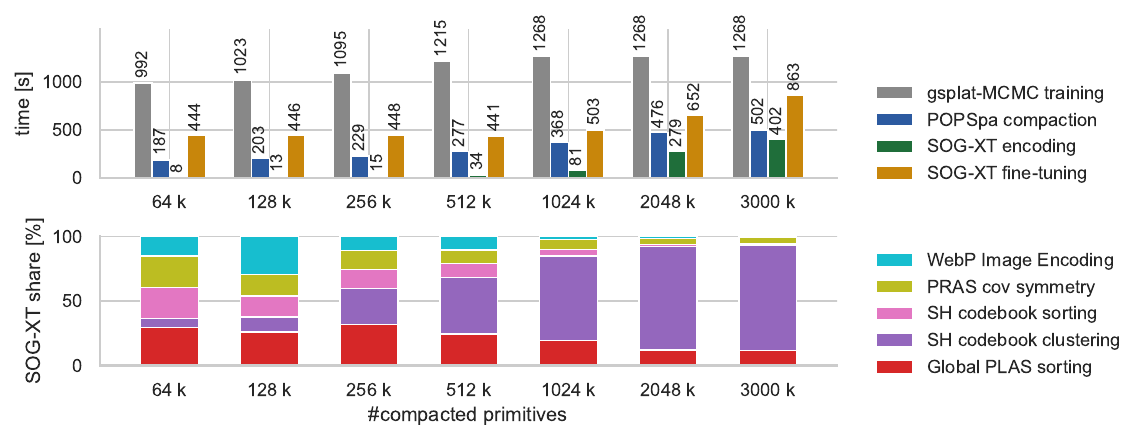}
  \caption{
  \textbf{Runtime breakdown} on the Bicycle scene.
  The top plot shows wall-clock time per KISS-GS stage and the bottom plot breaks down post-hoc SOG-XT encoding.
  Bicycle belongs to \MipNeRF, and 512k is the smallest Bicycle point exceeding INRIA-40k PSNR.
  At that point, reconstruction takes 1215 s, POPSpa compaction takes 277 s, post-hoc SOG-XT encoding takes 34 s, and optional encoding-aware fine-tuning adds 441 s.
  For context, Fig.~1 reports stage-wise size reductions of 15.7x, 6.6x, and 2.2x on \MipNeRF at INRIA-Q.
  Encoding costs are balanced between components for small scenes, while SH codebook clustering dominates at high primitive counts.
}
\label{fig:bicycle_encoding_time}
\end{figure}

\Cref{fig:bicycle_encoding_time} shows wall time for \textbf{encoding} with our KISS-GS pipeline.
Timings were measured using 16 AMD EPYC 7200 CPU cores and one Nvidia A100 GPU.
When comparing the wall time of different stages of the KISS-GS pipeline in the top view, we show that encoding with the SOG-XT format is inexpensive compared with reconstruction and compaction.
The optional encoding-aware fine-tuning is more costly because the codec is evaluated in the training loop.
In deployment, choosing between post-hoc and fine-tuned encoding is a trade-off between encoding time and bitrate, with the latter being more expensive but yielding better compression.
Both variants decode identically, so the adaptation improves bitrate without increasing deployment complexity.
Within post-hoc SOG-XT encoding (bottom view), the individual components have similar cost for small models. For larger models, SH codebook clustering starts to dominate and is the clearest target for future encoder optimization.

For \textbf{decoding} our format, we benchmark a reference Python decoder that reconstructs \ourCompression tensors using only standard WebP image decoding plus lightweight dequantization and table lookups.
On a Threadripper PRO 5955WX CPU, our end-to-end decoder takes 108 ms for 256k Gaussians, 192 ms for 512k, and 347 ms for 1024k.
These timings include file reads but exclude PLY writing.
Peak memory usage is 0.37 GB, 0.67 GB, and 1.24 GB, respectively.
The timing values are medians over the seven kept runs after one warmup run, with the two slowest of nine recorded runs discarded.
These measurements are intended as a conservative baseline because image decoding could be parallelized and handed off to hardware decoders, while per-field processing could be done more efficiently on the GPU.
They demonstrate that this unoptimized code can decode large 3DGS models in well under a second on commodity hardware, and that the memory footprint is small enough to fit comfortably in GPU memory for real-time rendering.

\subsection{Encoding-aware Fine-tuning Details}

In all experiments with encoding-aware fine-tuning, we run \num{4000} steps with a batch size of $8$.
Our implementation is based on PyTorch and uses gsplat for rendering.
We set a global step offset of \num{40}\,\text{k} to continue the learning-rate schedule after the reconstruction and compaction stages, which strongly reduces updates to the 3D positions.
The discrete structure of the codec is refreshed once at the start, including padding and the \emph{active mask}, PLAS ordering, the SH codebook assignments with $K \approx N/8$ clamped to $[\num{64}, \num{65536}]$, and PLAS sorting of the codebook.
We align the grid side length to a multiple of $8$ and use a PLAS minimum block size of $8$.
Quantization follows \ourCompression, including signed log for means with a 16-bit proxy split into byte planes, $q=99$ for quaternions, and 100-level quantization for SH codebook centroids.
Quantization ranges are computed once at initialization and kept fixed during training, except after \covopt{} updates.
We optimize a weighted combination of $\ell_1$ and SSIM with $\lambda_\mathrm{SSIM}=0.2$.
We add masked $\ell_1$ total variation on opacities, scales, both mean byte planes, and $f_\mathrm{dc}$, and we use a masked quaternion TV loss.
We set the TV weights to $1\mathrm{e}{-2}$ for quaternions and $1\mathrm{e}{-4}$ for all other terms.
These auxiliary losses are linearly warmed up from $0.35$ to $0.60$ of the run, corresponding to steps \num{1400} to \num{2400}.
We run \covopt{} once at initialization and again at $0.6$ and $0.8$ of the run, with PLAS as the driver.

\subsection{Encoding-aware Fine-tuning Sensitivity}

\begin{figure}[!htb]
  \centering
  \includegraphics[width=0.72\linewidth]{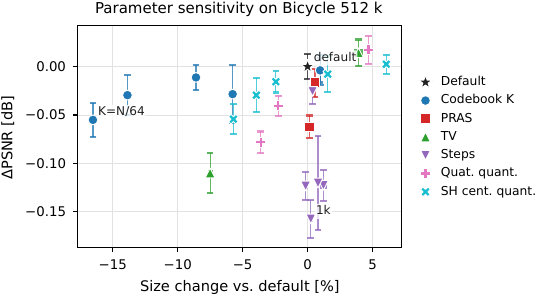}
  \caption{
  Parameter sensitivity of encoding-aware fine-tuning on the Bicycle scene with 512k Gaussians.
  We vary one parameter at a time around the default setting used in the paper:
  SH codebook size ($K\in N/\{2,4,12,16,32,64\}$ compared with the default $N/8$), \covopt{} timing, TV regularization (off, stronger), fine-tuning steps from 0.5k to 12k, quaternion quantization range 47--255, and SH-centroid quantization range 31--255.
  Points are means over seven runs, with error bars showing one standard deviation.
  }
  \label{fig:bicycle_sensitivity}
\end{figure}

\Cref{fig:bicycle_sensitivity} varies one parameter at a time around the default setting used in the paper.
The sweep shows a stable rate-distortion region rather than a fragile optimum on the Bicycle scene.
Disabling TV regularization or increasing quaternion precision gives only small PSNR gains while increasing file size.
Stronger TV regularization, very short fine-tuning, and additional fine-tuning steps do not improve the trade-off.
Changing the \covopt{} schedule stays close to or below the default, so we keep the fixed schedule described above.
The SH codebook size is the clearest optional rate-control knob, but its best setting is scene dependent.
We therefore use $K=N/8$ as a conservative default and use POPSpa primitive count as the primary rate-distortion control.

\subsection{Compaction Ablation}
\begin{table*}[!htb]

\caption{Ablation on the components of our compaction method, \ourCompaction. Adding the components individually, the results improve for two out of three of the real-world datasets and stagnate for the synthetic data.}

\begin{adjustbox}{width=12.2cm,center}
\rowcolors{2}{gray!25}{white}

\begin{tabular}{l|lllr|lllr|lllr|lllr}
\toprule
Method  & \multicolumn{4}{c|}{Tanks and Temples} & \multicolumn{4}{c|}{Mip-NeRF 360} & \multicolumn{4}{c|}{Deep Blending} & \multicolumn{4}{c|}{Synthetic NeRF} \\
 & \tiny PSNR$\uparrow$ & \tiny SSIM$\uparrow$ & \tiny LPIPS$\downarrow$ & \tiny k Gauss & \tiny PSNR$\uparrow$ & \tiny SSIM$\uparrow$ & \tiny LPIPS$\downarrow$ & \tiny k Gauss & \tiny PSNR$\uparrow$ & \tiny SSIM$\uparrow$ & \tiny LPIPS$\downarrow$ & \tiny k Gauss & \tiny PSNR$\uparrow$ & \tiny SSIM$\uparrow$ & \tiny LPIPS$\downarrow$ & \tiny k Gauss \\
\midrule
gsplat-MCMC~\cite{ye2025gsplat}~\venuetag{arXiv~'24} (30k)& 24.74 & .875 & .134 & 2,560 & 27.95 & .829 & .196 & 2,560 & 29.97 & .907 & .239 & 2,048 & 33.98 & .972 & .027 & 640 \\
\midrule
+ GaussianPOP~\cite{lee2026gaussianpop} (10k)& \cellcolor{lightorange}24.68 & \cellcolor{lightyellow}.864 & \cellcolor{lightyellow}.162 & \cellcolor{lightred}512 & \cellcolor{lightyellow}27.69 & \cellcolor{lightyellow}.802 & \cellcolor{lightyellow}.249 & \cellcolor{lightred}512 & \cellcolor{lightred}29.92 & \cellcolor{lightred}.906 & \cellcolor{lightorange}.249 & \cellcolor{lightred}512 & \cellcolor{lightred}33.83 & \cellcolor{lightorange}.970 & \cellcolor{lightred}.029 & \cellcolor{lightred}128 \\
+ Sparsification    & \cellcolor{lightred}24.77 & \cellcolor{lightorange}.870 & \cellcolor{lightorange}.149 & \cellcolor{lightred}512 & \cellcolor{lightorange}27.78 & \cellcolor{lightorange}.809 & \cellcolor{lightorange}.240 & \cellcolor{lightred}512 & \cellcolor{lightorange}29.85 & \cellcolor{lightorange}.905 & \cellcolor{lightred}.248 & \cellcolor{lightred}512 & \cellcolor{lightred}33.83 & \cellcolor{lightred}.971 & \cellcolor{lightred}.029 & \cellcolor{lightred}128 \\
+ Effective Rank Regularization    & \cellcolor{lightred}24.77 & \cellcolor{lightred}.871 & \cellcolor{lightred}.148 & \cellcolor{lightred}512 & \cellcolor{lightred}27.79 & \cellcolor{lightred}.811 & \cellcolor{lightred}.239 & \cellcolor{lightred}512 & \cellcolor{lightyellow}29.84 & \cellcolor{lightred}.906 & \cellcolor{lightyellow}.250 & \cellcolor{lightred}512 & \cellcolor{lightorange}33.73 & \cellcolor{lightorange}.970 & \cellcolor{lightred}.029 & \cellcolor{lightred}128 \\
\bottomrule
\end{tabular}
\end{adjustbox}
\label{tab:compaction_ablation}

\end{table*}

\Cref{tab:compaction_ablation} presents an ablation study of our compaction method.
Starting with a scene trained with gsplat-MCMC for 30k iterations, we apply two rounds of prune-refine using GaussianPOP's score.
Adding Sparsification to the first refinement stage generally improves the results, with the exception of the \DeepBlending dataset, which shows a slightly lower PSNR.
After adding the Effective Rank Regularization, our compaction still achieves competitive results, indicating that preparing the compacted data for compression introduces no quality loss.
Although this regularization is not intended to improve quality, it effectively reduces the spikiness of the Gaussians without degrading performance.

\subsection{Covariance Symmetries}
\begin{figure}[!htb]
  \centering
  \includegraphics[width=\linewidth]{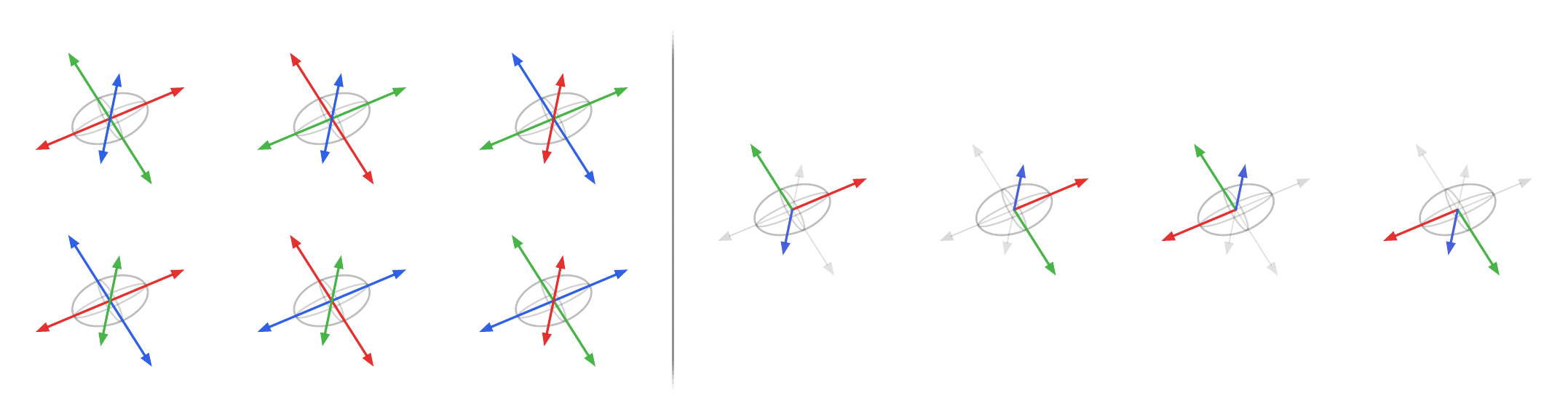}
  \caption{\textbf{Covariance symmetries}} \textbf{Left}: Scale permutations (6). Each permutation is indicated by a unique assignment of the local coordinate axes (red, green, and blue) to the principal axes of the ellipsoid. \\ \textbf{Right}: Valid eigenvector orientations (4) for a fixed scale permutation.
  \label{fig:covsym}
\end{figure}
\Cref{fig:covsym} illustrates the discrete symmetries inherent in the decomposition of the 3D covariance matrix $\Sigma = RSS^TR^T$, where $S=\text{diag}(s_1,s_2,s_3)$ are the scales and $R \in SO(3)$ are the rotation matrices.
As discussed in the main paper in Sec.~\ref{sec:encoding}, these symmetries give rise to $48$ equivalent parameterizations of the same Gaussian covariance.
As shown on the left, there are $3! = 6$ possible scale permutations, each defined by a unique assignment of the local coordinate axes to the ellipsoid's principal axes.
For any fixed permutation, we further identify four valid eigenvector orientations on the right.
While there are $2^3 = 8$ possible sign configurations for the eigenvectors, only four satisfy the $SO(3)$ constraint $\det(R) = 1$.
Together with the quaternion double cover, this yields the full set of $48$ equivalent choices used by \covopt.
\covopt exploits this freedom by evaluating the candidates within each equivalence class and selecting the one that improves the compressibility of the quaternion and scale grids without changing the represented covariance.

\subsection{HAC++ Results}

\begin{table*}[!htb]
\centering
\caption{Scores from the original HAC++ paper~\cite{chen2025hac-plus} and our own runs on \MipNeRF.
We list the reported 30k numbers, then our recomputed results with training resolution equal to evaluation resolution, and finally our recomputed results under the full evaluation protocol with 1600 px test resolution.
Scene size decreases up to 40k iterations and stabilizes thereafter, indicating convergence.}
\label{tab:hacpp_ablation}
\begin{tabular}{l|rrrr|rrrr}
\toprule
Method & \multicolumn{8}{c}{Mip-NeRF 360, train: 2x/4x, test: matched train resolution.} \\
\cmidrule{2-9}
 & \multicolumn{4}{c|}{Low-rate} & \multicolumn{4}{c}{High-rate} \\
 & \tiny PSNR$\uparrow$ & \tiny SSIM$\uparrow$ & \tiny LPIPS$\downarrow$ & \tiny Size & \tiny PSNR$\uparrow$ & \tiny SSIM$\uparrow$ & \tiny LPIPS$\downarrow$ & \tiny Size \\
\midrule
\rowcolor{white}
HAC++ (30k, reported)   & 27.60 & 0.803 & 0.253 & 8.34 & 27.82 & 0.811 & 0.231 & 18.48 \\
\midrule
\rowcolor{gray!25}
HAC++ (30k, recomputed) & 27.55 & 0.805 & 0.247 & 8.84 & 27.80 & 0.814 & 0.225 & 19.80 \\
\rowcolor{white}
HAC++ (40k, recomputed) & 27.33 & 0.802 & 0.250 & 8.41 & 27.80 & 0.813 & 0.225 & 19.68 \\
\rowcolor{gray!25}
HAC++ (44k, recomputed) & 27.20 & 0.800 & 0.251 & 8.42 & 27.79 & 0.813 & 0.226 & 19.67 \\
\midrule
\midrule
 & \multicolumn{8}{c}{Mip-NeRF 360, train: 2x/4x, test: 1600 px.} \\
\midrule
\rowcolor{gray!25}
HAC++ (30k, recomputed) & 27.43 & 0.800 & 0.256 & 8.82 & 27.57 & 0.806 & 0.235 & 19.76 \\
\rowcolor{white}
HAC++ (40k, recomputed) & 27.21 & 0.796 & 0.260 & 8.38 & 27.62 & 0.806 & 0.236 & 19.62 \\
\rowcolor{gray!25}
HAC++ (44k, recomputed) & 27.09 & 0.795 & 0.261 & 8.39 & 27.61 & 0.806 & 0.236 & 19.62 \\
\bottomrule
\end{tabular}
\end{table*}

\Cref{tab:hacpp_ablation} shows the original HAC++ numbers alongside our own runs on \MipNeRF.
We first list the reported 30k results from the HAC++ paper.
This 30k setting is also the default training budget for 3DGS, which makes it the natural reference point.
We then report our recomputed 30k, 40k, and 44k runs with training resolution equal to evaluation resolution.
Finally, we report the same runs under our full evaluation protocol from \cref{sec:eval_protocol}, including 1600 px test resolution.
We include 44k because our own pipeline adds 10k steps in the compaction stage and 4k steps in the fine-tuning stage, so this gives HAC++ the same overall compute budget in this comparison.
Our 30k recomputation in the matched train and eval setting is close to the reported values.
Moving to the full evaluation protocol lowers the quality metrics slightly, while file size remains nearly unchanged.
Extending the compute budget to 40k and 44k iterations demonstrates size convergence: scene size reduces slightly at 40k in the low-rate regime, but additional steps do not improve rendering quality.
This is consistent with the HAC++ loss:
\begin{equation}
    \mathcal{L} = \mathcal{L}_{\text{Scaffold}} + \lambda \frac{1}{N(D_a + 6 + 3K)}\left(\mathcal{L}_{\text{entropy}} + \mathcal{L}_{\text{hash}}\right)
\end{equation}
where $\lambda$ controls the rate-distortion trade-off. At high $\lambda$ (low-rate), extended 
training forces further compression without recovering quality. At low $\lambda$ (high-rate), 
this pressure is relaxed and quality remains stable regardless of iteration count.

\subsection{Evaluation Protocol}
\label{sec:eval_protocol}
For our evaluation experiments, we follow the evaluation protocol established in the codebase of the original 3DGS paper\cite{kerbl3Dgaussians}.
The 3DGS.zip survey\cite{3DGSzip2025} states the scene selection, the training and evaluation resolutions, and the training and test split.
Here, we add details on LPIPS and the background color.
\begin{itemize}
\item 21 scenes from four datasets (two real-world scenes in \TandT, a mix of nine indoor and outdoor scenes in \MipNeRF, two in \DeepBlending and eight synthetic scenes in the \SyntheticNeRF dataset) show a range of tiny synthetic scenes to medium-large real-world scenes.
\item Training resolution is 4x downsampled for \MipNeRF outdoor scenes, 2x downsampled for \MipNeRF indoor scenes, and otherwise 1x the provided images. 
\item Evaluation uses images downsampled to 1600 px on the longest side rather than the training resolution.
\item Image downsampling is handled with PIL bicubic resampling.
\item The background color is white for \SyntheticNeRF, and black otherwise.
\item Metrics are PSNR, SSIM, and LPIPS, where for LPIPS the VGG backbone is used, and the images are not normalized to $[-1, 1]$ but are instead kept in their $[0, 1]$ range\cite{kulhanek2025nerfbaselines}.
\item Metrics are evaluated on a hold-out validation set, which comprises every 8th image of the real-world datasets and the designated test split of \SyntheticNeRF.
\end{itemize}

Because metrics are computed on held-out images, this protocol evaluates \emph{novel view generalization} rather than fidelity on the training set alone. This distinction is particularly important in the context of compression, where a compact representation should preserve not only the appearance of the observed views but also the scene structure required to render unseen viewpoints accurately. The evaluation thus helps ensure that bitrate reductions are not achieved by compressing away the generalization capacity.

In practice, the evaluation-resolution mismatch affects only \MipNeRF, as the remaining datasets have image resolutions below 1600 pixels on the longest side. \MipNeRF is therefore the most informative benchmark in this protocol: it covers a diverse set of real-world indoor and outdoor scenes, contains multiple scenes rather than isolated examples, and jointly tests novel view generalization and robustness to differing training and evaluation resolutions.

\subsection{Compression Ablation}

Additional ablations of our compression format \ourCompression are reported in \cref{tab:compression-ablation-full}. Across nearly all attributes, uniform min–max quantization is highly effective: values are stored using a small fixed number of discrete levels, while the original minimum and maximum are retained as metadata for dequantization. Depending on the attribute, we use \num{100}, \num{256}, or \num{65536} quantization levels, denoted in the table as \texttt{q100\_u8}, \texttt{u8}, and \texttt{u16}, respectively. This reduces size substantially while causing only minor quality loss.

\begin{table*}[!htb]

\caption{Ablation of contributions to the \ourCompression format on the Bicycle scene with \num{256}\,\text{k} primitives.
The first row contains the absolute values after encoding. The following rows disable individual features of \ourCompression, ordered by size penalty. Most removals increase the final size considerably at small quality gains. Using image encoding, our novel codebook sorting and its UV label packing bring clean size wins without incurring any quality loss.
Compared with the main paper version, additional rows included only in the appendix are marked by a \colorbox{blue!25}{colored bar} in the first column.}

\begin{adjustbox}{width=10.5cm,center}
\rowcolors{1}{gray!25}{white}
\newcommand{\newrowmark}{\cellcolor{blue!25}}
\newcolumntype{L}{>{\centering\arraybackslash}p{1mm}}

\small
\setlength{\tabcolsep}{4pt}
\small
\setlength{\tabcolsep}{4pt}
\begin{tabular}{L p{5.5cm} r r r r r}
\hiderowcolors %
& & & \multicolumn{4}{c}{\textbf{Effect of Leaving Feature Out}} \\
\cmidrule(lr){4-7}
& & Size [MB] & \makecell[r]{$\Delta$ Size [Bytes] $\downarrow$} & \makecell[r]{$\Delta$ PSNR $\uparrow$} & \makecell[r]{$\Delta$ SSIM $\uparrow$} & \makecell[r]{$\Delta$ LPIPS $\downarrow$} \\
\cmidrule(lr){2-7}
& \textbf{SOG-XT (Full Pipeline)} & 3.878 & (3,877,944) & (24.2281) & (0.68176) & (0.35404) \\
\midrule
\showrowcolors %
\noalign{\global\rownum=1} %
\newrowmark & w/o centroid q100\_u8 quantization & 7.920 & \textcolor{DiffRed}{+4,042,523} & \textcolor{DiffGreen}{+0.0097} & \textcolor{DiffGreen}{+0.00016} & \textcolor{DiffGreen}{-0.00032} \\
& w/o SH AC codebooks & 7.313 & \textcolor{DiffRed}{+3,434,816} & \textcolor{DiffGreen}{+0.2320} & \textcolor{DiffGreen}{+0.00790} & \textcolor{DiffGreen}{-0.00579} \\
& w/o WebP image compression & 6.908 & \textcolor{DiffRed}{+3,030,148} & \textcolor{DiffGray}{+0.0000} & \textcolor{DiffGray}{+0.00000} & \textcolor{DiffGray}{+0.00000} \\
& w/o quaternion u8 quantization & 6.427 & \textcolor{DiffRed}{+2,548,718} & \textcolor{DiffGreen}{+0.0317} & \textcolor{DiffGreen}{+0.00147} & \textcolor{DiffGreen}{-0.00067} \\
\newrowmark & w/o f\_dc u8 quantization & 5.962 & \textcolor{DiffRed}{+2,084,291} & \textcolor{DiffGreen}{+0.0063} & \textcolor{DiffGreen}{+0.00012} & \textcolor{DiffGreen}{-0.00009} \\
\newrowmark & w/o scales u8 quantization & 5.844 & \textcolor{DiffRed}{+1,966,415} & \textcolor{DiffGreen}{+0.0040} & \textcolor{DiffGreen}{+0.00023} & \textcolor{DiffGreen}{-0.00013} \\
\newrowmark & w/o means u16 quantization & 5.271 & \textcolor{DiffRed}{+1,392,843} & \textcolor{DiffGreen}{+0.0121} & \textcolor{DiffGreen}{+0.00049} & \textcolor{DiffGreen}{-0.00020} \\
& w/o primitive PLAS sorting & 4.493 & \textcolor{DiffRed}{+614,851} & \textcolor{DiffGreen}{+0.0059} & \textcolor{DiffGreen}{+0.00022} & \textcolor{DiffGreen}{-0.00036} \\
\newrowmark & w/o opacities u8 quantization & 4.487 & \textcolor{DiffRed}{+609,117} & \textcolor{DiffGreen}{+0.0006} & \textcolor{DiffGreen}{+0.00007} & \textcolor{DiffGreen}{-0.00006} \\
\newrowmark & w/o \covopt, quaternion full u8 quant & 4.040 & \textcolor{DiffRed}{+161,748} & \textcolor{DiffGreen}{+0.0101} & \textcolor{DiffGreen}{+0.00040} & \textcolor{DiffGreen}{-0.00008} \\
& w/o centroid PLAS sorting & 3.920 & \textcolor{DiffRed}{+42,416} & \textcolor{DiffGray}{+0.0000} & \textcolor{DiffGray}{+0.00000} & \textcolor{DiffGray}{+0.00000} \\
& w/o label UV packing & 3.890 & \textcolor{DiffRed}{+12,142} & \textcolor{DiffGray}{+0.0000} & \textcolor{DiffGray}{+0.00000} & \textcolor{DiffGray}{+0.00000} \\
\newrowmark & w/o grid mask (prune to square) & 3.881 & \textcolor{DiffRed}{+3,344} & \textcolor{DiffRed}{-0.1150} & \textcolor{DiffRed}{-0.00576} & \textcolor{DiffRed}{+0.00279} \\
& w/o \covopt & 3.866 & \textcolor{DiffGreen}{-11,810} & \textcolor{DiffRed}{-0.1104} & \textcolor{DiffRed}{-0.00538} & \textcolor{DiffRed}{+0.00273} \\
& w/o means signed-log remapping & 3.359 & \textcolor{DiffGreen}{-519,272} & \textcolor{DiffRed}{-7.5508} & \textcolor{DiffRed}{-0.37864} & \textcolor{DiffRed}{+0.22151} \\
\end{tabular}
\end{adjustbox}

\label{tab:compression-ablation-full}

\end{table*}

For quaternions, \ourCompression uses particularly aggressive min–max quantization with only \num{100} levels, referred to as \texttt{q100\_u8} in the table. This is enabled by \covopt: when \covopt is removed, the same quantization becomes too coarse and causes a pronounced quality drop. Recovering the quality requires relaxing the quantization to a denser \num{256}-level representation, denoted as \texttt{u8}, as shown in the row “w/o \covopt, quaternion full u8 quant”.

PLAS sorting requires the primitives to fit into square 2D grids. In overcomplete representations, this can be handled by pruning surplus primitives, for example, by removing the lowest-opacity ones as in SOG\cite{morgenstern2024compact}. After \ourCompaction, however, the remaining Gaussians are already valuable, making further pruning undesirable. \ourCompression therefore pads the grid with dummy primitives and uses an active mask to discard them again at decode time. Replacing this with opacity-based pruning provides only a tiny size benefit but leads to a clear quality loss ("w/o grid mask (prune to square)").

\subsection{Simple Decoding of the \ourCompression Format}

In the paper, we argue that although the encoding pipeline is sophisticated, decoding can remain simple.
To illustrate this point, we provide a compact Python reference implementation that decodes one of our output files and writes the reconstructed scene to an INRIA-style `.ply` file.

The script first defines a small set of helpers for reading quantization ranges from the metadata, dequantizing stored attributes, and unpacking the tiled spherical harmonics centroid table into the 45 AC components.

It then decodes the attribute images, applies the corresponding inverse transforms, reconstructs the per-primitive attributes, and writes the result as a `.ply` file.

\begin{lstlisting}[style=pyreadable]
#!/usr/bin/env -S uv run --script
# /// script
# requires-python = ">=3.12"
# dependencies = [
#     "numpy",
#     "pillow",
#     "pyyaml",
# ]
# ///

"""Decode a SOG-XT container directory into a 3DGS-INRIA `.ply`."""

from __future__ import annotations

import argparse
from pathlib import Path

import numpy as np
import yaml
from PIL import Image


# ============================================================================
# Imports and Helpers
# ============================================================================

def read_image_uint8(path: Path) -> np.ndarray:
    image = np.array(Image.open(path))
    if image.ndim == 2:
        image = image[..., None]
    return image.astype(np.uint8, copy=False)


def normalize_to_01(values: np.ndarray) -> np.ndarray:
    values_f32 = values.astype(np.float32, copy=False)
    return (values_f32 - values_f32.min()) / (values_f32.max() - values_f32.min() + 1e-8)


def dequantize(
    quantized: np.ndarray,
    min_values: np.ndarray | float,
    max_values: np.ndarray | float,
) -> np.ndarray:
    return normalize_to_01(quantized) * (max_values - min_values) + min_values


def extract_ranges(container_meta: dict) -> dict[str, tuple[object, object]]:
    ranges: dict[str, tuple[object, object]] = {}
    for op in container_meta["ops"]:
        fields = op.get("input_fields") or []
        if len(fields) != 1:
            continue
        for transform in op.get("transforms", []):
            quant = transform.get("simple_quantize")
            if quant and "min_values" in quant:
                ranges[fields[0]] = (quant["min_values"], quant["max_values"])
    return ranges


def untile_feature_grid_3x5(tiled_hw3: np.ndarray) -> np.ndarray:
    tile_rows, tile_cols = 3, 5
    total_height, total_width, channels = tiled_hw3.shape
    height = total_height // tile_rows
    width = total_width // tile_cols
    tiles = tiled_hw3.reshape(tile_rows, height, tile_cols, width, channels)
    return tiles.transpose(1, 3, 4, 0, 2).reshape(height, width, channels * tile_rows * tile_cols)


# ============================================================================
# Main Decoding Path
# ============================================================================

def decode_container_to_columns(container_dir: Path) -> list[tuple[str, np.ndarray]]:
    container_meta = yaml.safe_load((container_dir / "container_meta.yaml").read_text())
    grid_side = int(container_meta["meta"]["grid_side"])
    centroid_grid_side = int(container_meta["meta"]["f_rest_centroids_side"])
    image_codec = str(container_meta["meta"]["image_codec"])
    ranges = extract_ranges(container_meta)

    def p(field: str) -> Path:
        return container_dir / f"{field}.{image_codec}"

    active_mask = read_image_uint8(p("active_mask"))[..., 0].reshape(-1) > 0

    opacities = dequantize(
        read_image_uint8(p("opacities"))[..., 0],
        float(ranges["opacities"][0]),
        float(ranges["opacities"][1]),
    )
    opacities = (1.0 / (1.0 + np.exp(-opacities))).reshape(grid_side * grid_side, 1)

    scales_min = np.asarray(ranges["scales"][0], dtype=np.float32).reshape(1, 1, 3)
    scales_max = np.asarray(ranges["scales"][1], dtype=np.float32).reshape(1, 1, 3)
    scales = np.exp(dequantize(read_image_uint8(p("scales")), scales_min, scales_max)).reshape(grid_side * grid_side, 3)

    means_low = read_image_uint8(p("means_bytes_0")).astype(np.float32)
    means_high = read_image_uint8(p("means_bytes_1")).astype(np.float32)
    means_signed_log = dequantize(
        means_low + 256.0 * means_high,
        float(ranges["means"][0]),
        float(ranges["means"][1]),
    )
    means = (np.sign(means_signed_log) * np.expm1(np.abs(means_signed_log))).reshape(grid_side * grid_side, 3)

    quaternions_min = np.asarray(ranges["quaternions"][0], dtype=np.float32).reshape(1, 1, 4)
    quaternions_max = np.asarray(ranges["quaternions"][1], dtype=np.float32).reshape(1, 1, 4)
    quaternions = dequantize(
        read_image_uint8(p("quaternions")),
        quaternions_min,
        quaternions_max,
    ).reshape(grid_side * grid_side, 4)

    f_dc_min = np.asarray(ranges["f_dc"][0], dtype=np.float32).reshape(1, 1, 3)
    f_dc_max = np.asarray(ranges["f_dc"][1], dtype=np.float32).reshape(1, 1, 3)
    f_dc = dequantize(read_image_uint8(p("f_dc")), f_dc_min, f_dc_max).reshape(grid_side * grid_side, 3)

    labels_uv = read_image_uint8(p("f_rest_labels")).astype(np.int64, copy=False)
    centroid_indices = (labels_uv[..., 1] * centroid_grid_side + labels_uv[..., 0]).reshape(-1)

    centroid_min = np.asarray(ranges["f_rest_centroids"][0], dtype=np.float32)
    centroid_max = np.asarray(ranges["f_rest_centroids"][1], dtype=np.float32)
    if centroid_min.size != 45:
        raise ValueError("Expected 45-channel centroids (SOG-XT).")

    centroid_grid = untile_feature_grid_3x5(normalize_to_01(read_image_uint8(p("f_rest_centroids"))))
    centroids = dequantize(
        centroid_grid.reshape(centroid_grid_side, centroid_grid_side, 45),
        centroid_min.reshape(1, 1, 45),
        centroid_max.reshape(1, 1, 45),
    ).reshape(centroid_grid_side * centroid_grid_side, 45)

    f_rest = centroids[centroid_indices].reshape(grid_side * grid_side, 3, 15).transpose(0, 2, 1)
    sh = np.concatenate([f_dc[:, None, :], f_rest], axis=1)

    means = means[active_mask]
    scales = scales[active_mask]
    opacities = opacities[active_mask]
    quaternions = quaternions[active_mask]
    sh = sh[active_mask]

    normals = np.zeros_like(means, dtype=np.float32)
    scales_log = np.log(scales)
    opacities_logit = np.log(opacities) - np.log1p(-opacities)
    f_rest_flat = sh[:, 1:, :].transpose(0, 2, 1).reshape(sh.shape[0], 45)

    return [
        ("x", means[:, 0]),
        ("y", means[:, 1]),
        ("z", means[:, 2]),
        ("nx", normals[:, 0]),
        ("ny", normals[:, 1]),
        ("nz", normals[:, 2]),
        ("f_dc_0", sh[:, 0, 0]),
        ("f_dc_1", sh[:, 0, 1]),
        ("f_dc_2", sh[:, 0, 2]),
        *[(f"f_rest_{i}", f_rest_flat[:, i]) for i in range(45)],
        ("opacity", opacities_logit[:, 0]),
        ("scale_0", scales_log[:, 0]),
        ("scale_1", scales_log[:, 1]),
        ("scale_2", scales_log[:, 2]),
        ("rot_0", quaternions[:, 0]),
        ("rot_1", quaternions[:, 1]),
        ("rot_2", quaternions[:, 2]),
        ("rot_3", quaternions[:, 3]),
    ]


# ============================================================================
# PLY Writer and CLI Entry Point
# ============================================================================

def write_ply(path: Path, *, columns: list[tuple[str, np.ndarray]]) -> None:
    path.parent.mkdir(parents=True, exist_ok=True)
    num_vertices = int(columns[0][1].shape[0])
    matrix = np.empty((num_vertices, len(columns)), dtype="<f4")
    for column_index, (_name, values) in enumerate(columns):
        matrix[:, column_index] = values.astype(np.float32, copy=False).reshape(num_vertices)

    header = "\n".join([
        "ply",
        "format binary_little_endian 1.0",
        f"element vertex {num_vertices}",
        *[f"property float {name}" for name, _ in columns],
        "end_header",
        "",
    ]).encode("ascii")

    with path.open("wb") as f:
        f.write(header)
        f.write(matrix.tobytes())


def main() -> None:
    parser = argparse.ArgumentParser(description=__doc__)
    parser.add_argument("--input", type=Path, required=True, help="Input container directory.")
    parser.add_argument("--output", type=Path, required=True, help="Output INRIA `.ply` path.")
    args = parser.parse_args()
    write_ply(args.output, columns=decode_container_to_columns(args.input))


if __name__ == "__main__":
    main()
\end{lstlisting}

\subsection{Comparison with State of the Art}

\begin{figure}[!htb]
  \centering
  \includegraphics[width=\linewidth]{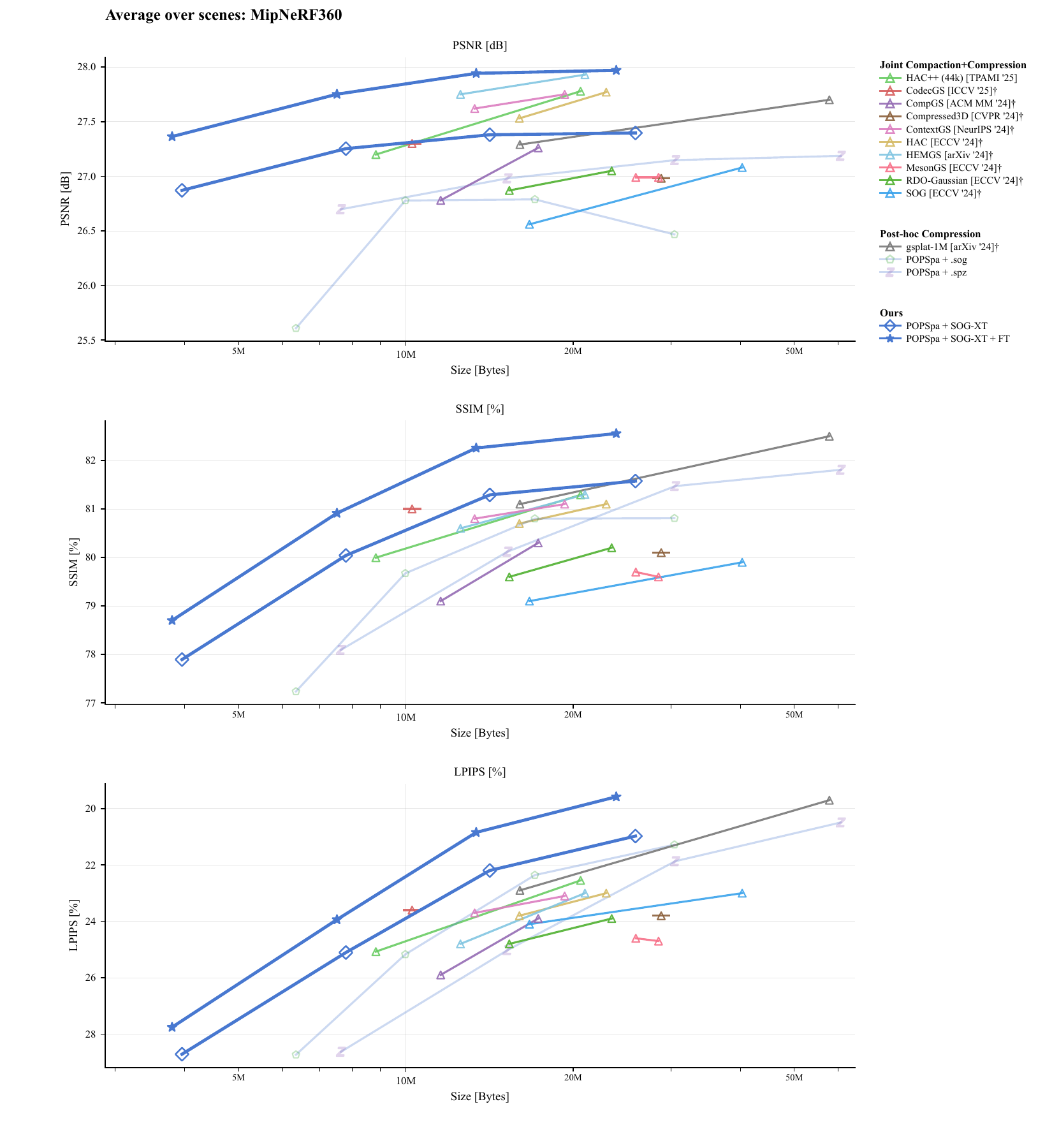}
  \caption{
  Rate-distortion performance of KISS-GS and related work on the \MipNeRF scenes.
  Methods marked with a dagger \textsuperscript{$\dagger$} are self-reported by the authors, as collected by the 3DGS.zip survey\cite{3DGSzip2025}, and may not follow our evaluation protocol from \cref{sec:eval_protocol}.
}
  \label{fig:grouped-rd-mipnerf-comparison}
\end{figure}

\Cref{fig:grouped-rd-mipnerf-comparison} places our \MipNeRF results in the broader context of recent 3DGS compression methods from the literature.
\MipNeRF is the most informative benchmark in our evaluation protocol because it contains multiple real-world scenes, is evaluated on held-out views, and exposes the effect of using different training and evaluation resolutions.
Even with this broader comparison, our reproduced operating points remain competitive across the rate-distortion range, with particularly strong results in LPIPS and file size.

\subsection{Rate-Distortion Curves}

The full curves make the dataset differences more explicit.
On \TandT and \MipNeRF, our method benefits from a clear quality margin over INRIA after reconstruction and compaction, which can then be traded for bitrate during compression.
\DeepBlending is the exception, where this margin is largely absent.
The gap to HAC++ is already present before SOG-XT encoding, since compressed HAC++ exceeds uncompressed gsplat-MCMC and INRIA 3DGS on both \textit{Playroom} and \textit{Dr.~Johnson}.
This points to the underlying representation and its regularization rather than to SOG-XT quantization or POPSpa pruning.
Our per-view inspection indicates that the mean gap is driven by a small number of views where high-capacity standard splats form isolated floaters.
Reducing the primitive budget with POPSpa removes many of these weakly constrained primitives, which explains why the 512k models remain close to HAC++ on the full scenes.
We therefore view \DeepBlending as a limitation of the standard-splat reconstruction used as input to KISS-GS under these captures, rather than as a limitation of the image-based encoder alone.

\subsection{Qualitative Comparison}

\Cref{tab:view-comp-bicycle-17} complements the quantitative tables and rate-distortion curves with a view-level comparison on the Bicycle scene.
The examples support the same trend as the quantitative results.
At 128k primitives and only 1.9 MB, our method already reconstructs the main foreground structure well and still reaches competitive PSNR, but it loses fine background detail.
This can be seen in the grass behind the pedals in the 4x crop, the gray label on the bicycle frame on the left in the 8x crop, and the background behind the wheels in the 12x crop.
At 1024k primitives, most of this missing background structure is recovered while the file remains slightly smaller than HAC++ low-rate, which still misses these details at a similar size.
At 2048k primitives, the reconstruction also surpasses the INRIA baseline in LPIPS while remaining far smaller on disk, at 25.0 MB versus 1334.9 MB.

\clearpage

\subsection{Full-Page Figures and Tables}

For convenience, we collect the large comparison figures and tables here in the same order in which they were discussed above.

\begin{figure}[!htb]
  \centering
  \includegraphics[width=\linewidth]{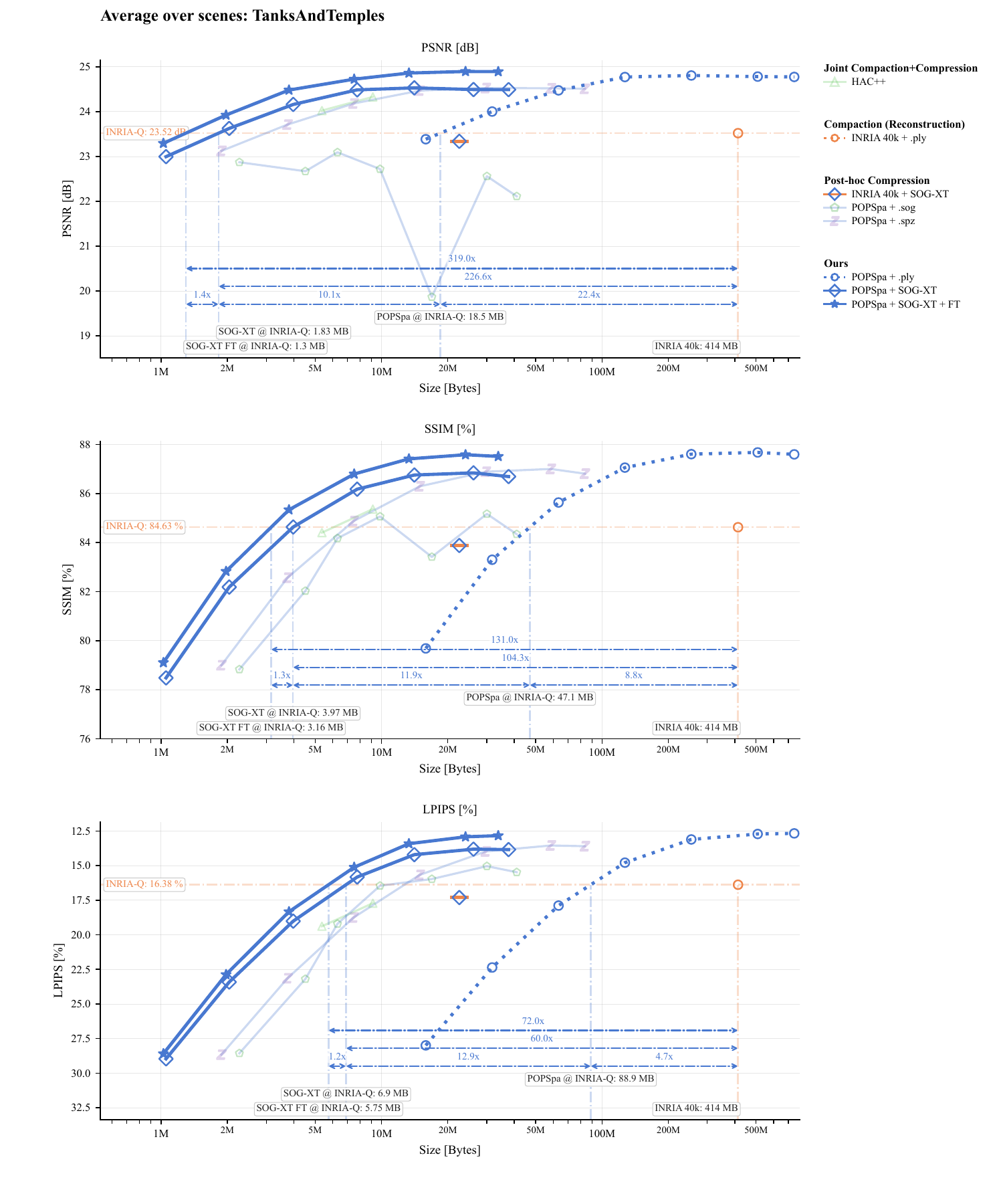}
  \caption{
  Rate-distortion performance of KISS-GS on the \TandT scenes.
  Our reconstruction and compaction pipeline substantially surpasses the INRIA 40k baseline, which creates headroom for compression.
  As a result, KISS-GS can match the INRIA baseline at file sizes as low as $1.3$ MB in PSNR and $5.75$ MB in LPIPS.
}
  \label{fig:grouped-rd-tandt}
\end{figure}

\begin{figure}[!htb]
  \centering
  \includegraphics[width=\linewidth]{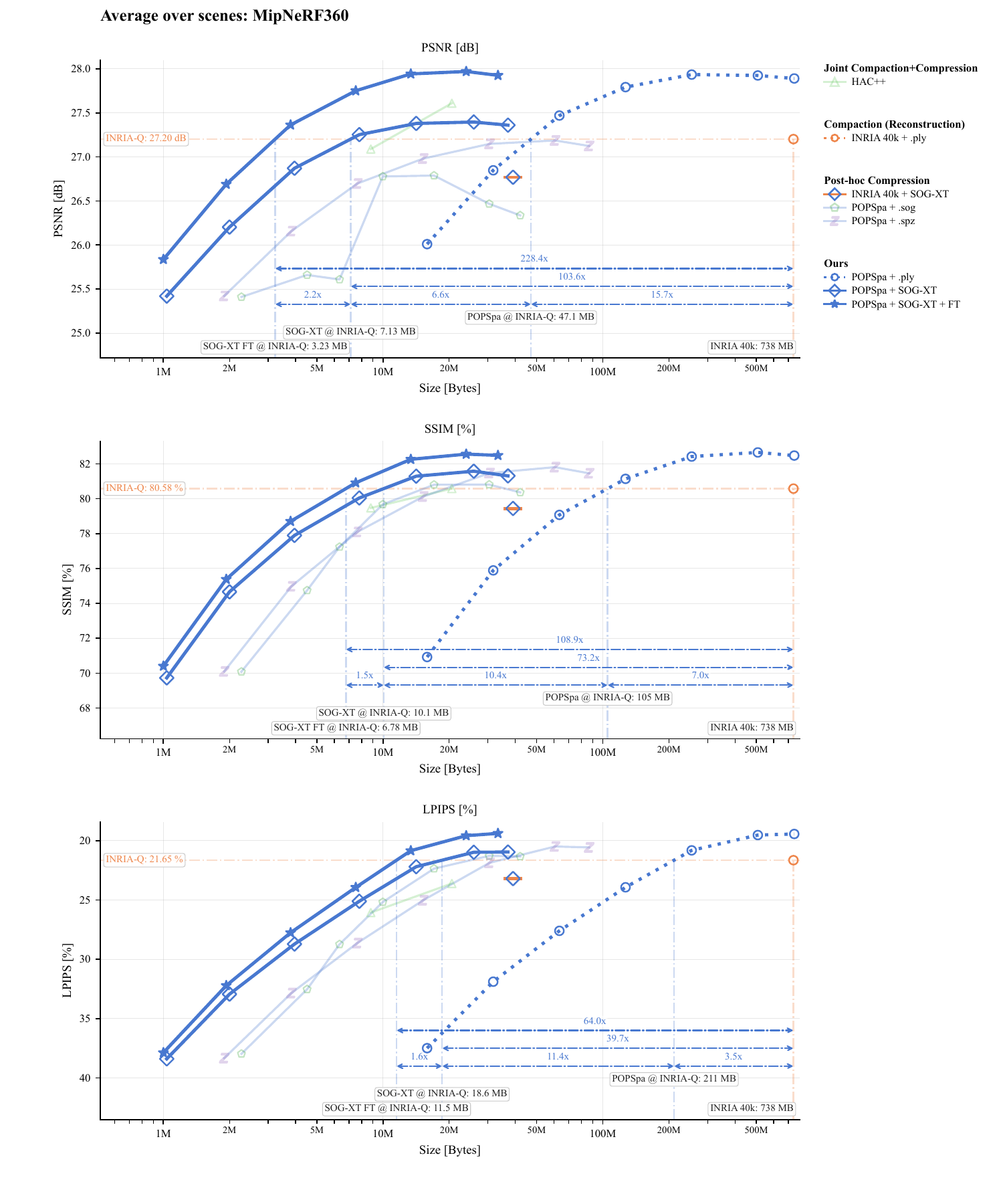}
  \caption{
  Rate-distortion performance of KISS-GS on the \MipNeRF scenes:
  In addition to the PSNR plot in \cref{fig:main_rd}, we report SSIM and LPIPS.
  The ordering is similar across metrics, although the gaps are closer for SSIM and LPIPS than for PSNR.
}
  \label{fig:grouped-rd-mipnerf}
\end{figure}

\begin{figure}[!htb]
  \centering
  \includegraphics[width=\linewidth]{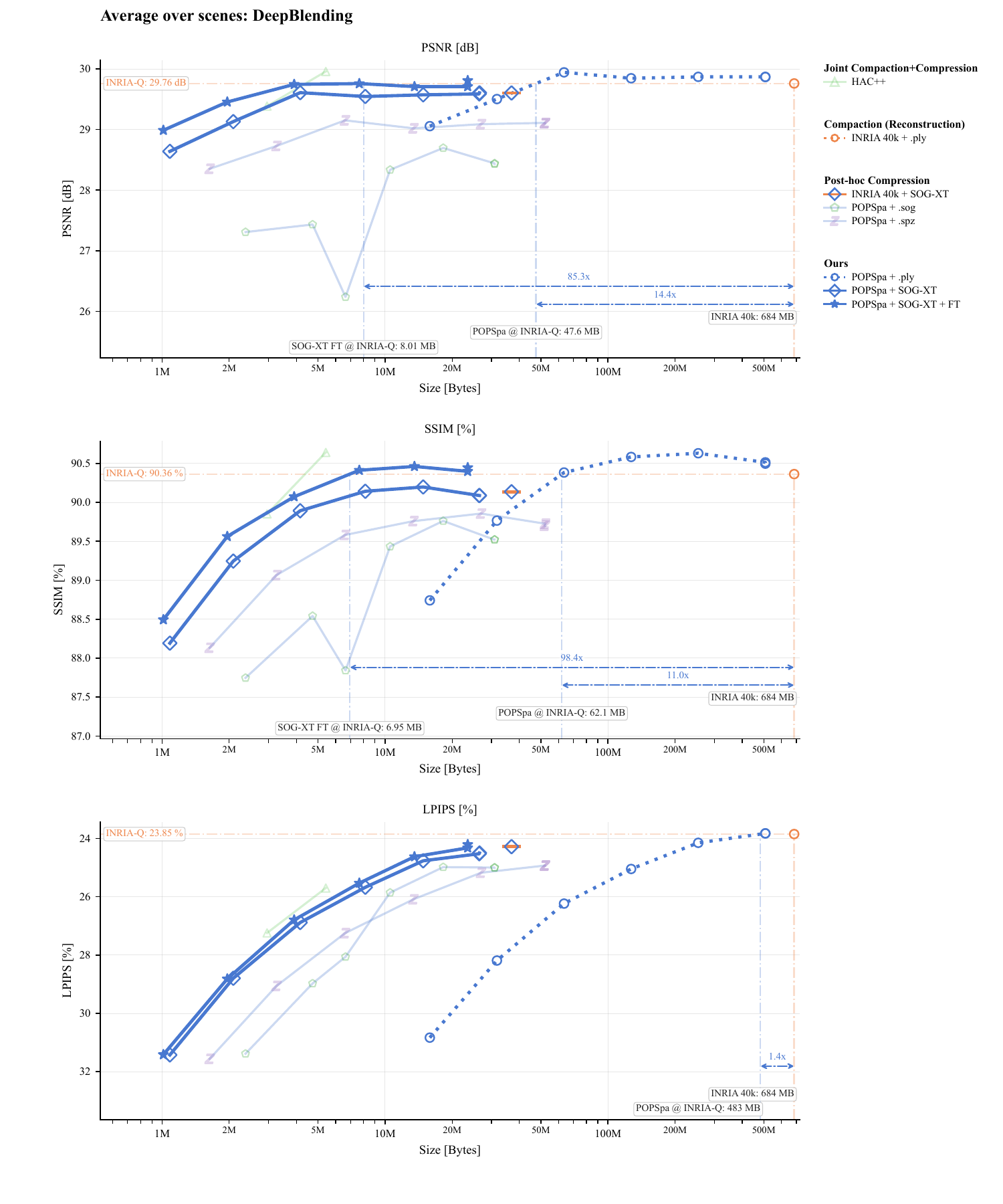}
  \caption{
  Rate-distortion performance of KISS-GS on the \DeepBlending scenes.
  Unlike on the other datasets, our reconstruction and compaction pipeline does not substantially improve over the INRIA baseline.
  The gap to HAC++ is already present before SOG-XT encoding, which indicates that the limiting factor is the standard-splat reconstruction under these captures rather than the image-based compression stage alone.
  As a result, none of the operating points of our compressed curve exceeds INRIA quality, while HAC++\cite{chen2025hac-plus} remains above INRIA for part of the range.
}
  \label{fig:grouped-rd-deepblending}
\end{figure}

\begin{figure}[!htb]
  \centering
  \includegraphics[width=\linewidth]{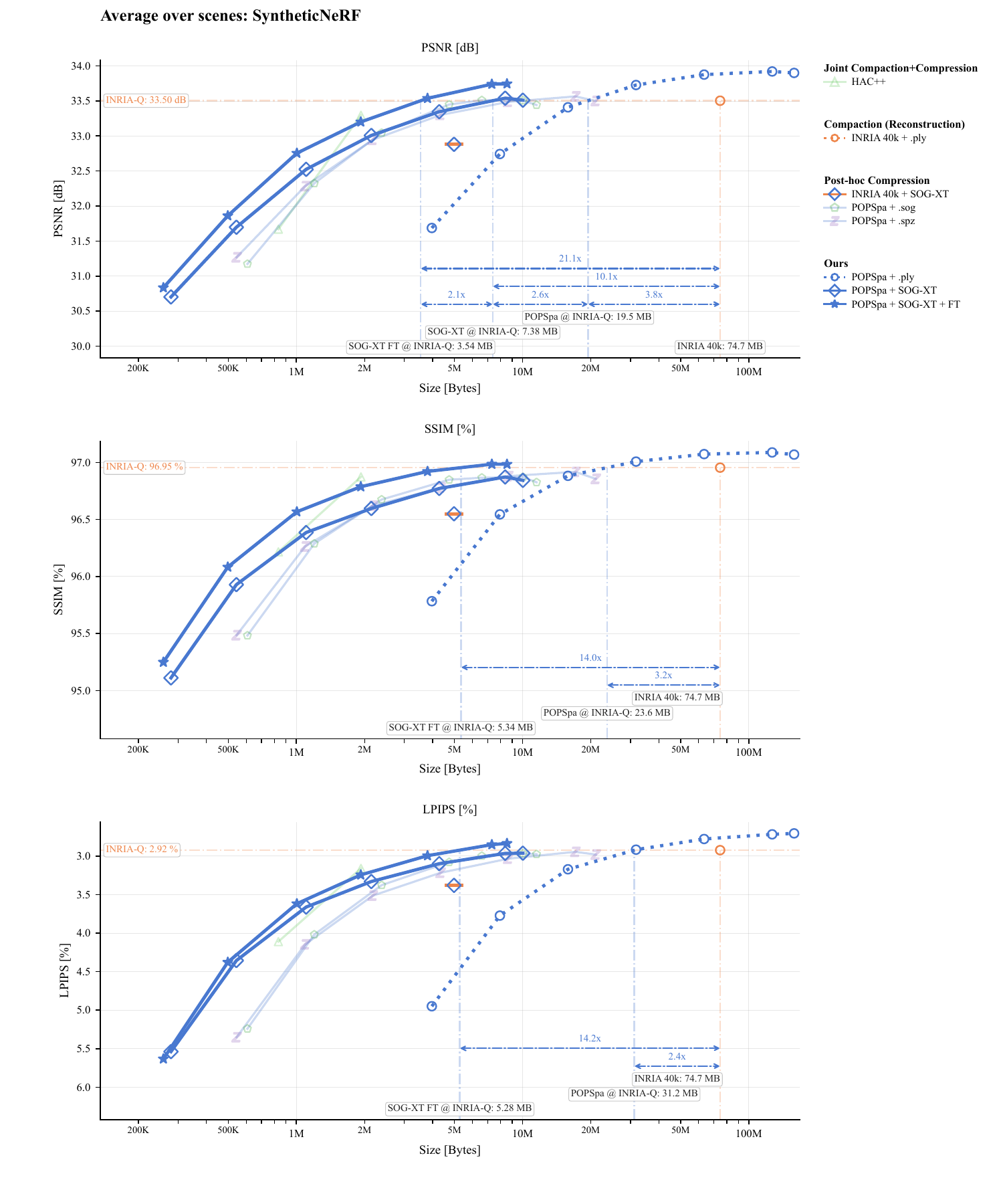}
  \caption{
  Rate-distortion performance of KISS-GS on the \SyntheticNeRF scenes.
  Because these scenes are synthetic, spatially smaller, and free of real-world capture effects, absolute quality metrics are higher for all methods.
  }
  \label{fig:grouped-rd-syntheticnerf}
\end{figure}

\newcommand{\VCMethodColW}{1.8cm}
\newcommand{\VCMetricColW}{1.9cm}

\begin{table*}[!htb]
\centering
\caption{Qualitative comparison on evaluation view 17 of the Bicycle scene.
We compare INRIA 40k, HAC++-44k, and our method (\ourCompaction + \ourCompression + Finetuning) at five primitive budgets.
The Metrics column reports PSNR, LPIPS, and file size for this view.
Informative regions to inspect are the grass behind the pedals in the 4x crop, the gray label on the bicycle frame on the left in the 8x crop, and the background behind the wheels in the 12x crop.
}
\label{tab:view-comp-bicycle-17}
\begingroup
\scriptsize
\begin{tblr}{
width=\linewidth,
rows = {valign=t},
rowsep = 5pt,
colsep = 4pt,
stretch = 0,
vspan = even,
colspec = {Q[l,wd=\VCMethodColW] Q[l,wd=\VCMetricColW] X[c] X[c] X[c] X[c]},
row{1} = {font=\bfseries,abovesep=2pt,belowsep=3pt},
}
\hline[1pt]
Method & Metrics & 1x & 4x & 8x & 12x \\
\hline[0.8pt]
Ground truth &  & \SetCell[r=3]{c,m} \includegraphics[width=\linewidth]{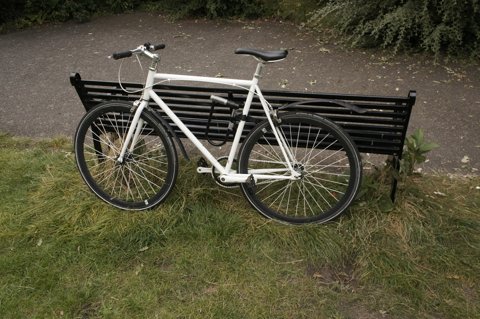} & \SetCell[r=3]{c,m} \includegraphics[width=\linewidth]{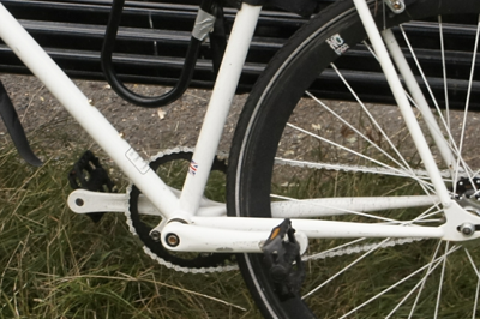} & \SetCell[r=3]{c,m} \includegraphics[width=\linewidth]{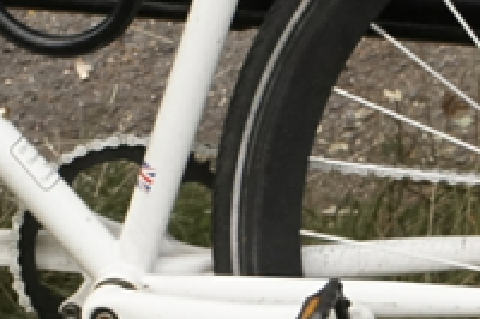} & \SetCell[r=3]{c,m} \includegraphics[width=\linewidth]{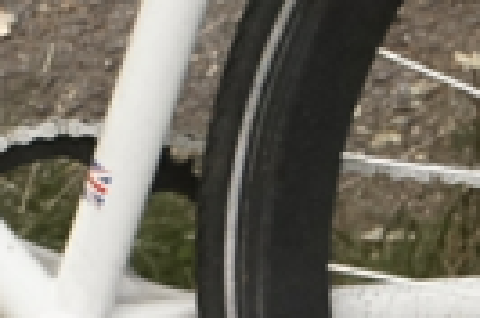} \\
 &  &  &  &  &  \\
 &  &  &  &  &  \\
\hline[0.6pt]
INRIA & PSNR 24.806 & \SetCell[r=3]{c,m} \includegraphics[width=\linewidth]{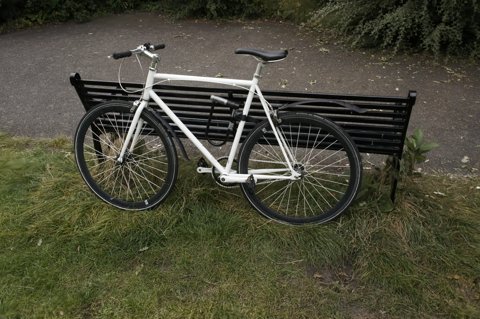} & \SetCell[r=3]{c,m} \includegraphics[width=\linewidth]{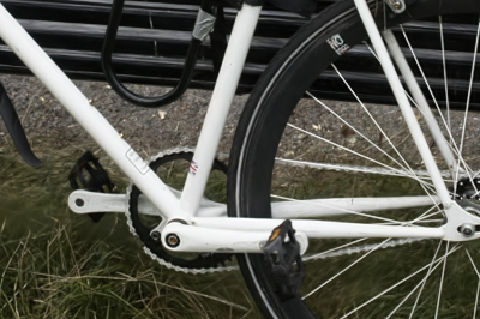} & \SetCell[r=3]{c,m} \includegraphics[width=\linewidth]{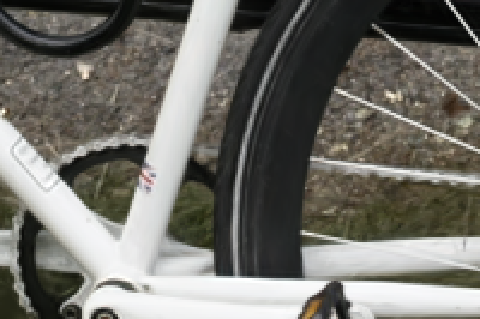} & \SetCell[r=3]{c,m} \includegraphics[width=\linewidth]{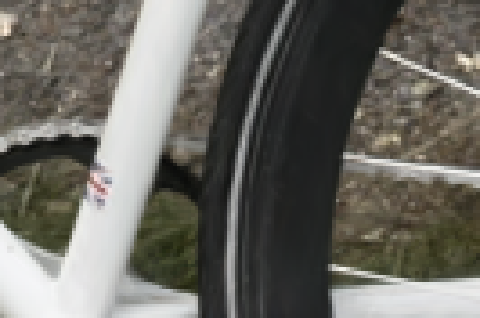} \\
40k & LPIPS 0.1662 &  &  &  &  \\
 & Size 1334.9 MB &  &  &  &  \\
\hline[0.6pt]
HAC++ & PSNR 25.223 & \SetCell[r=3]{c,m} \includegraphics[width=\linewidth]{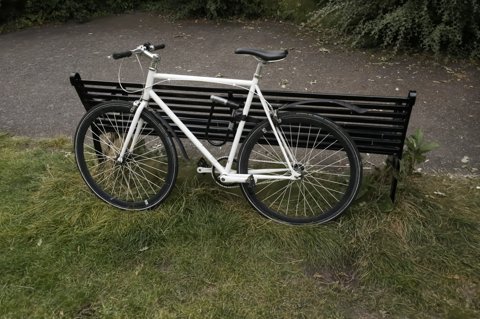} & \SetCell[r=3]{c,m} \includegraphics[width=\linewidth]{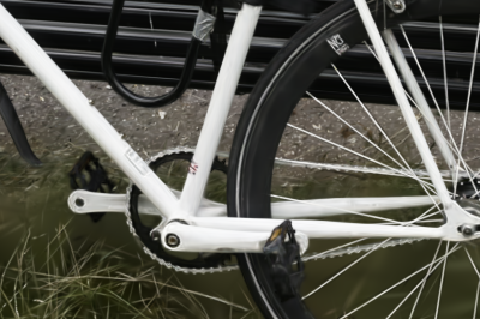} & \SetCell[r=3]{c,m} \includegraphics[width=\linewidth]{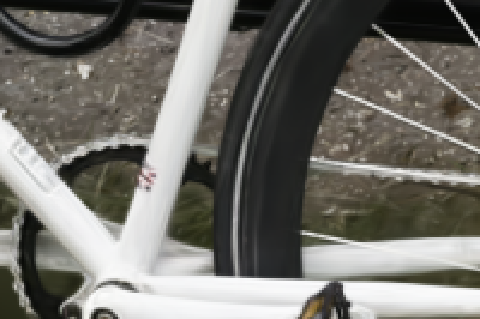} & \SetCell[r=3]{c,m} \includegraphics[width=\linewidth]{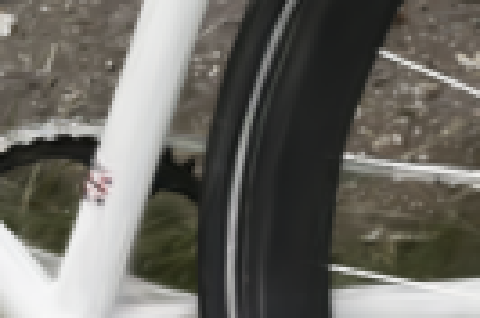} \\
low\mbox{-}rate & LPIPS 0.2533 &  &  &  &  \\
 & Size 13.8 MB &  &  &  &  \\
\hline[0.6pt]
HAC++ & PSNR 25.023 & \SetCell[r=3]{c,m} \includegraphics[width=\linewidth]{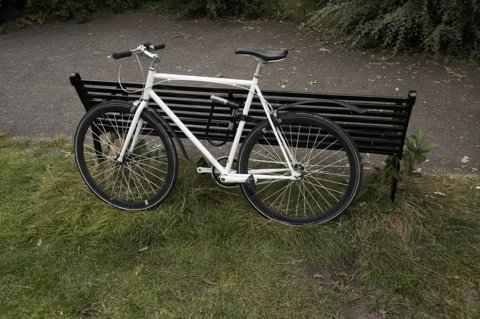} & \SetCell[r=3]{c,m} \includegraphics[width=\linewidth]{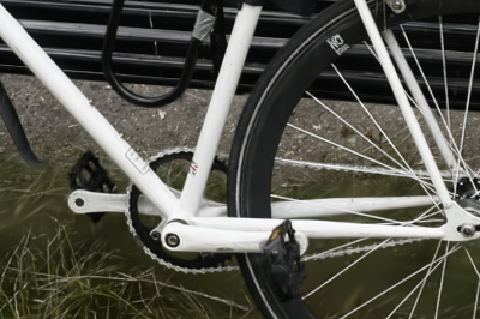} & \SetCell[r=3]{c,m} \includegraphics[width=\linewidth]{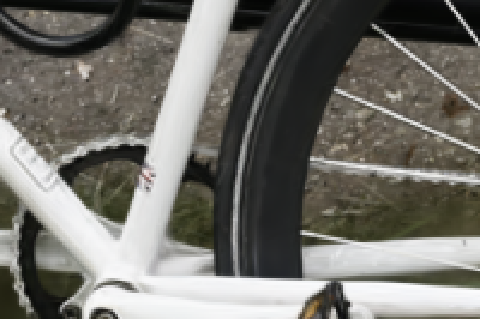} & \SetCell[r=3]{c,m} \includegraphics[width=\linewidth]{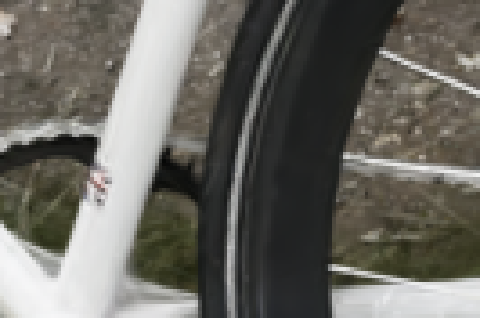} \\
high\mbox{-}rate & LPIPS 0.2174 &  &  &  &  \\
 & Size 33.5 MB &  &  &  &  \\
\hline[0.6pt]
POPSpa\mbox{-}128k & PSNR 24.949 & \SetCell[r=3]{c,m} \includegraphics[width=\linewidth]{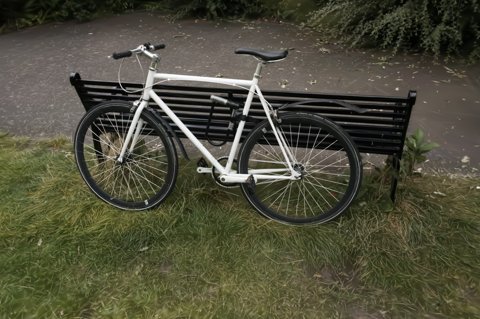} & \SetCell[r=3]{c,m} \includegraphics[width=\linewidth]{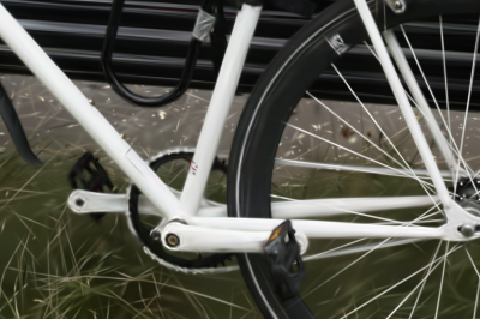} & \SetCell[r=3]{c,m} \includegraphics[width=\linewidth]{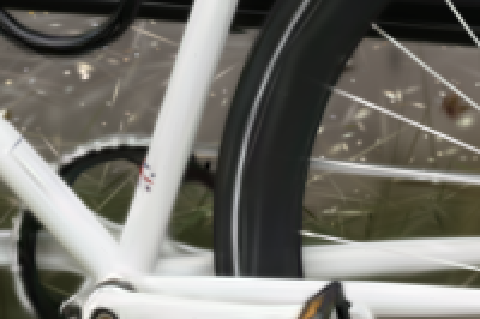} & \SetCell[r=3]{c,m} \includegraphics[width=\linewidth]{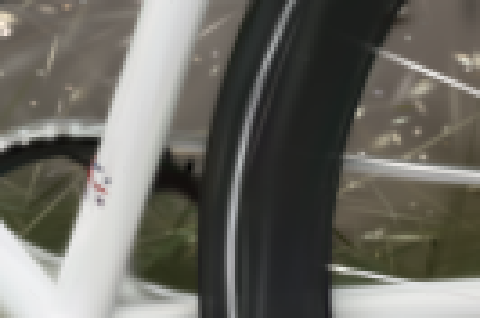} \\
SOG\mbox{-}XT\mbox{-}FT & LPIPS 0.3474 &  &  &  &  \\
(Ours) & Size 1.9 MB &  &  &  &  \\
\hline[0.6pt]
POPSpa\mbox{-}256k & PSNR 25.613 & \SetCell[r=3]{c,m} \includegraphics[width=\linewidth]{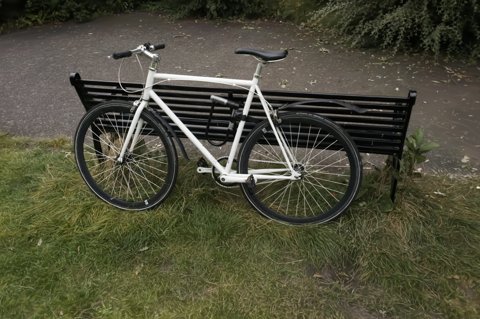} & \SetCell[r=3]{c,m} \includegraphics[width=\linewidth]{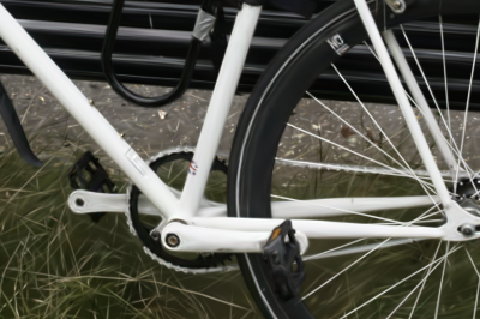} & \SetCell[r=3]{c,m} \includegraphics[width=\linewidth]{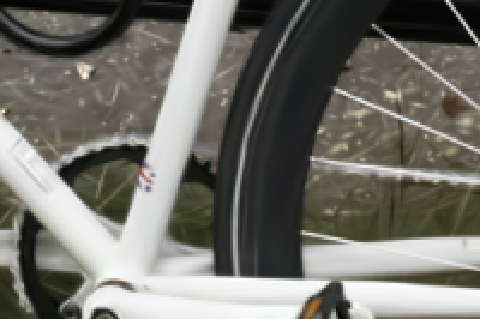} & \SetCell[r=3]{c,m} \includegraphics[width=\linewidth]{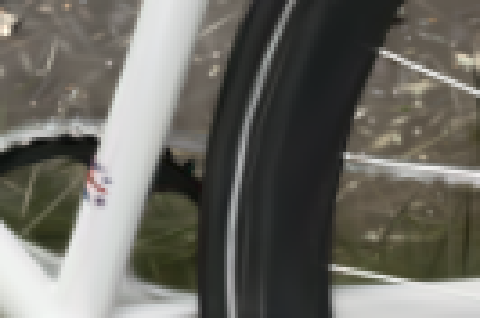} \\
SOG\mbox{-}XT\mbox{-}FT & LPIPS 0.2925 &  &  &  &  \\
(Ours) & Size 3.7 MB &  &  &  &  \\
\hline[0.6pt]
POPSpa\mbox{-}512k & PSNR 25.973 & \SetCell[r=3]{c,m} \includegraphics[width=\linewidth]{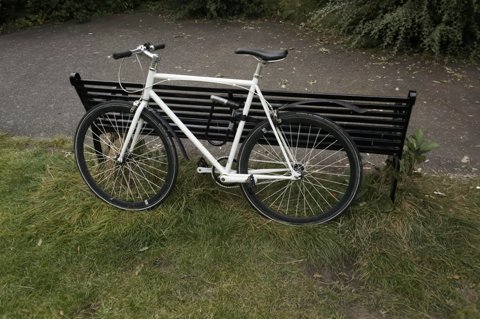} & \SetCell[r=3]{c,m} \includegraphics[width=\linewidth]{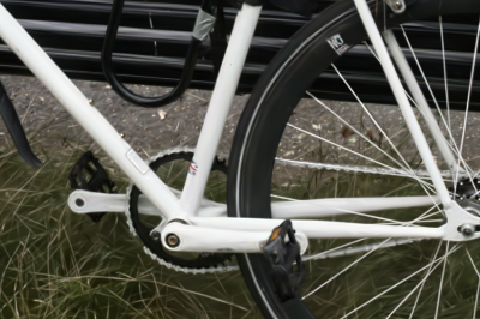} & \SetCell[r=3]{c,m} \includegraphics[width=\linewidth]{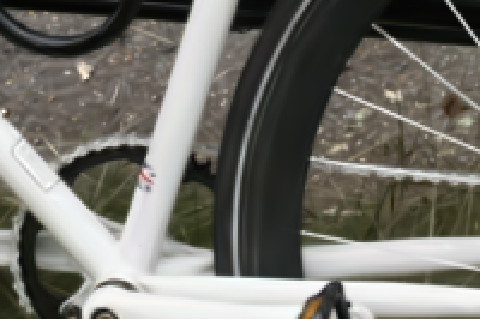} & \SetCell[r=3]{c,m} \includegraphics[width=\linewidth]{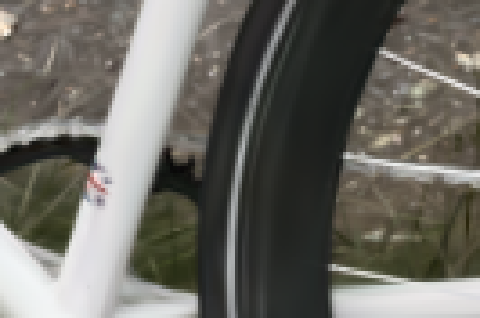} \\
SOG\mbox{-}XT\mbox{-}FT & LPIPS 0.2380 &  &  &  &  \\
(Ours) & Size 7.4 MB &  &  &  &  \\
\hline[0.6pt]
POPSpa\mbox{-}1024k & PSNR 25.990 & \SetCell[r=3]{c,m} \includegraphics[width=\linewidth]{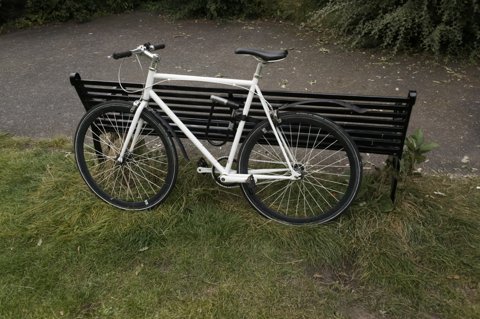} & \SetCell[r=3]{c,m} \includegraphics[width=\linewidth]{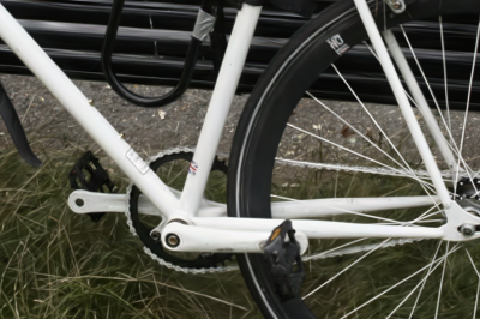} & \SetCell[r=3]{c,m} \includegraphics[width=\linewidth]{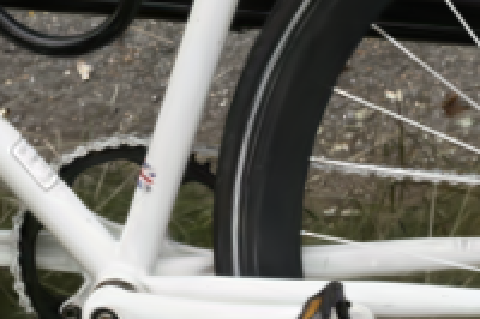} & \SetCell[r=3]{c,m} \includegraphics[width=\linewidth]{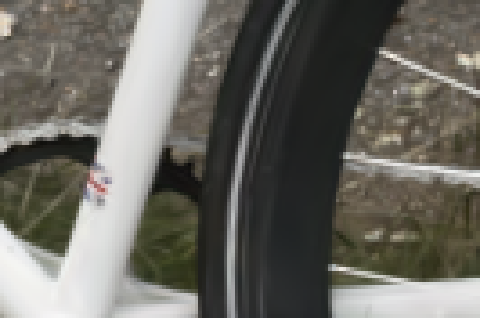} \\
SOG\mbox{-}XT\mbox{-}FT & LPIPS 0.1812 &  &  &  &  \\
(Ours) & Size 13.5 MB &  &  &  &  \\
\hline[0.6pt]
POPSpa\mbox{-}2048k & PSNR 25.663 & \SetCell[r=3]{c,m} \includegraphics[width=\linewidth]{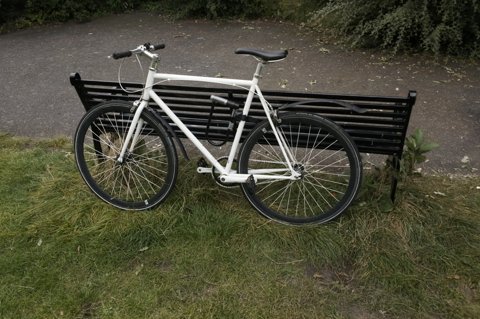} & \SetCell[r=3]{c,m} \includegraphics[width=\linewidth]{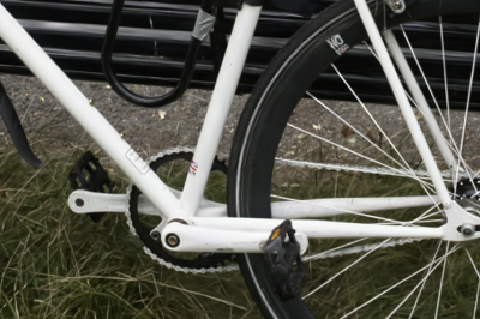} & \SetCell[r=3]{c,m} \includegraphics[width=\linewidth]{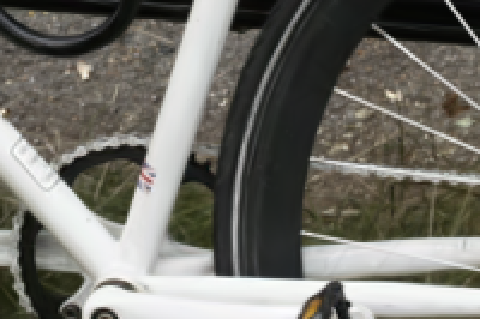} & \SetCell[r=3]{c,m} \includegraphics[width=\linewidth]{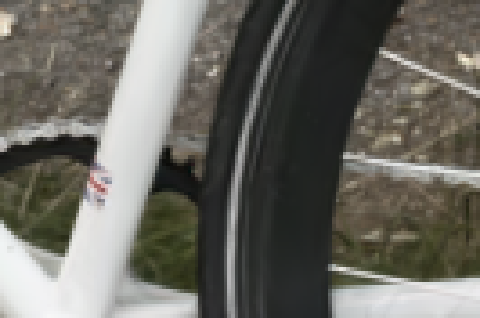} \\
SOG\mbox{-}XT\mbox{-}FT & LPIPS 0.1515 &  &  &  &  \\
(Ours) & Size 25.0 MB &  &  &  &  \\
\hline[1pt]
\end{tblr}
\endgroup
\end{table*}

\end{document}